\documentclass[twoside,11pt]{article}
\usepackage{jmlr2e}
\usepackage{lastpage}
\usepackage{amsmath,color}
\usepackage{graphicx,psfrag,epsf}
\usepackage{enumerate}
\usepackage{natbib}
\usepackage{url}
\usepackage{amssymb}
\usepackage{multirow}
\usepackage{array}
\usepackage{bm}
\usepackage{booktabs}
\usepackage{mathtools}
\usepackage{setspace}
\usepackage{algorithm,algorithmic}
\usepackage[font=small,labelfont=bf]{caption}

\newcommand{\Mean}{{\mbox{E}}}
\newcommand{\Var}{{\mbox{Var}}}
\newcommand{\Cov}{{\mbox{cov}}}

\newcommand{\prob}{{\mbox{Pr}}}

\def\eop{\hfill $\Box$}

\let\proglang=\textsf

\newcommand\independent{\protect\mathpalette{\protect\independenT}{\perp}}
\def\independenT#1#2{\mathrel{\rlap{$#1#2$}\mkern2mu{#1#2}}}

\newcommand{\change}[1]{{\leavevmode\color{black}{#1}}}

\begin{document}

\title{Double Generative Adversarial Networks for Conditional Independence Testing}	

\author{\name Chengchun Shi \email{c.shi7@lse.ac.uk} \\ \name Tianlin Xu \email{T.Xu12@lse.ac.uk}\\ 
\name Wicher Bergsma \email{w.p.bergsma@lse.ac.uk} \\
\addr Department of Statistics, London School of Economics and Political Science \\ 
\AND
\name Lexin Li \email{lexinli@berkeley.edu}\\
\addr Department of Biostatistics and Epidemiology, University of California at Berkeley} 

\editor{}

\maketitle

\begin{abstract}
In this article, we study the problem of high-dimensional conditional independence testing, a key building block in statistics and machine learning. We propose an inferential procedure based on double generative adversarial networks (GANs). Specifically, we first introduce a double GANs framework to learn two generators of the conditional distributions. We then integrate the two generators to construct a test statistic, which takes the form of the maximum of generalized covariance measures of multiple transformation functions. We also employ data-splitting and cross-fitting to minimize the conditions on the generators to achieve the desired asymptotic properties, and employ multiplier bootstrap to obtain the corresponding $p$-value. We show that the constructed test statistic is doubly robust, and the resulting test both controls type-I error and has the power approaching one asymptotically. Also notably, we establish those theoretical guarantees under much weaker and practically more feasible conditions compared to the existing tests, and our proposal gives a concrete example of how to utilize some state-of-the-art deep learning tools, such as GANs, to help address a classical but challenging statistical problem. We demonstrate the efficacy of our test through both simulations and an application to an anti-cancer drug dataset. A \proglang{Python} implementation of the proposed procedure is available at \url{https://github.com/tianlinxu312/dgcit}.
\end{abstract}
\bigskip


\section{Introduction}
\label{secintroduction}

Conditional independence (CI) is a fundamental concept in statistics and machine learning. Testing conditional independence is a key building block and plays a central role in a large variety of statistical learning problems, for instance, causal inference \citep{pearl2009}, graphical models \citep{koller2009}, dimension reduction \citep{lib2018}, among many others. It is frequently used in a wide range of scientific and business applications, and we demonstrate its application with a cancer genetics example later. 

In this article, we aim at testing whether two random variables $X$ and $Y$ are conditionally independent given a set of confounding variables $Z$. That is, we test the hypotheses:
\begin{eqnarray}\label{eqn:test}
\mathcal{H}_0: X \independent Y \; | \; Z \quad\quad \textrm{versus} \quad\quad \mathcal{H}_1: X \not\independent Y \; | \; Z,
\end{eqnarray}
given the observed data of $n$ i.i.d.\ copies $\{(X_i,Y_i,Z_i)\}_{1\le i\le n}$ of $(X,Y,Z)$. For our problem, $X, Y$ and $Z$ can all be multivariate. However, the main challenge arises when the confounding set of variables $Z$ is multivariate and high-dimensional. As such, we primarily focus on the scenario where $X$ and $Y$ are univariate, and $Z$ is multivariate and its dimension can potentially diverge to infinity. Meanwhile, our proposed method can be readily extended to the scenario of  multivariate $X$ and $Y$ as well. Another challenge is the limited sample size compared to the dimensionality of $Z$. As a result, many existing tests may become ineffective, suffering from either an inflated type-I error, or not having enough power to detect the alternatives. See Section \ref{sec:relatedwork} for a detailed literature review.  

To deal with those challenges, we propose a testing procedure based on double generative adversarial networks \citep[GANs,][]{goodfellow2014generative} for the CI testing problem in (\ref{eqn:test}). GANs have recently stood out as a powerful approach for learning and generating random samples from a complex, high-dimensional data distribution. They have been successfully applied in numerous applications, ranging from image processing and computer vision, to sequential data modeling such as natural language, music, speech, and to medical fields such as DNA design and drug discovery; see \cite{gui2020review} for a review of the GANs applications. Moreover, there have recently emerged works studying the consistency and rate of convergence of the GANs estimators; see, e.g., \citet{liang2018well,chen2020statistical}. 

Our proposal involves two key components: a double GANs framework to learn two generators that approximate the conditional distribution of $X$ given $Z$, and $Y$ given $Z$, respectively, and a test statistic that is taken as the maximum of generalized covariance measures of multiple transformation functions of $X$ and $Y$. We first show that our test statistic is doubly-robust, which offers an additional layer of protection against potential misspecification of the conditional distributions; see Theorems \ref{thm1} and \ref{thm2}. We then show that the resulting test achieves a valid control of the type-I error asymptotically, and more importantly, under the set of conditions that are much weaker and practically more feasible compare to the existing tests; see Theorem \ref{thm3}. Besides, we prove that the power of our test approaches one asymptotically; see Theorem \ref{thm4}, and we demonstrate through simulations that it is more powerful than numerous competing tests empirically. In addition, we employ data splitting and cross-fitting that allow us to derive the asymptotic properties under minimal conditions on the generators, and employ multiplier bootstrap to obtain the corresponding $p$-value of the test. Our contributions are multi-fold. We develop a useful testing procedure for a fundamentally important statistical inference problem. We establish the statistical guarantees under much weaker conditions. We also give an example of how to utilize some state-of-the-art deep learning tools, such as GANs, to address a classical but challenging statistical problem.

The rest of the article is organized as follows. Section \ref{sec:relatedwork} reviews some key existing CI testing methods. Section \ref{sec:doubleGAN} develops the double GANs-based testing procedure. Section \ref{sec:theory} derives the theoretical properties. Section \ref{sec:numeric} presents the simulations and a cancer genetics data example. Section \ref{sec:dis} concludes the paper. The Appendix collects all technical proofs.

\section{Literature review on conditional independence testing}
\label{sec:relatedwork}

There has been a large and growing literature on conditional independence testing; see \citet{li2019nonparametric} for a review. Broadly speaking, the existing tests can be cast into four main categories, the metric-based tests \citep[e.g.,][]{su2007consistent, su2014testing, wang2015conditional,pan2017conditional,wang2018characteristic}, the conditional randomization-based tests \citep[e.g.,][]{candes2018panning, bellot2019conditional}, the kernel-based tests \citep[e.g.,][]{fukumizu2008kernel, zhang2011kernel}, and the regression-based tests \citep[e.g.,][]{hoyer2009nonlinear, shah2018hardness}. There are also some other types of tests \citep[e.g.,][to name a few]{bergsma2004testing, berrett2019conditional}. 

The metric-based tests typically employ some kernel smoothers to estimate the conditional characteristic function or the distribution function of $Y$ given $X$ and $Z$. Kernel smoothers, however, are known to suffer from the curse of dimensionality, and as such, these tests are usually not suitable when the dimension of $Z$ is high. The conditional randomization-based tests require the knowledge of the conditional distribution of $X | Z$ \citep{candes2018panning}. If unknown, the type-I error rates of these tests rely critically on the quality of the approximation of this conditional distribution. Kernel-based tests are built upon the notion of maximum mean discrepancy \citep[MMD,][]{gretton2012kernel}, and could have inflated type-I errors. Regression-based tests have valid type-I error control, but may suffer from inadequate power. 

Next, we discuss in detail the conditional randomization-based tests, in particular, the work of \cite{bellot2019conditional}, the regression-based tests, and the MMD-based tests, as our proposal is related to and built on those methods. For each family of tests, we first lay out the main ideas, then discuss their potential limitations.

\subsection{Conditional randomization-based tests}
\label{secCRT}

The family of conditional randomization-based tests is built upon the following basis. If the conditional distribution $P_{X|Z}$ of $X$ given $Z$ is known, then one can independently draw $X_i^{(1)} \sim P_{X|Z=Z_i}$, for $i=1,\ldots,n$, where the superscript denotes the first round of draws. Besides, these samples are independent of the observed samples $X_i$'s and $Y_i$'s. Write $\bm{X} = (X_1,\ldots,X_n)^\top$, $\bm{X}^{(1)}=(X_1^{(1)},\ldots,X_n^{(1)})^\top$, $\bm{Y}=(Y_1,\ldots,Y_n)^\top$, and $\bm{Z}= (Z_1,\ldots,Z_n)^\top$. Hereinafter we use boldface letters to denote data matrices that consist of $n$ samples. Since the joint distributions of $(\bm{X},\bm{Y},\bm{Z})$ and $(\bm{X}^{(1)},\bm{Y},\bm{Z})$ are the same under $\mathcal{H}_0$, any large difference between the two distributions can be interpreted as evidence against $\mathcal{H}_0$. Therefore, one can repeat the sample drawing process $M$ times, i.e., $X_i^{(m)}\sim P_{X|Z=Z_i}$, $i=1,\ldots,n$,  $m=1,\ldots,M$. Write $\bm{X}^{(m)}=(X_1^{(m)},\ldots,X_n^{(m)})^\top$. Then, for a given test statistic $\rho=\rho(\bm{X},\bm{Y},\bm{Z})$, the associated $p$-value is 
\begin{eqnarray*}
p = \frac{1}{M} \left[\sum_{m=1}^{M} \mathbb{I} \left\{ \rho(\bm{X}^{(m)},\bm{Y},\bm{Z})\ge \rho(\bm{X},\bm{Y},\bm{Z}) \right\} \right] ,
\end{eqnarray*}
where $\mathbb{I}(\cdot)$ denotes the indicator function. Since the triplets $(\bm{X},\bm{Y},\bm{Z}), (\bm{X}^{(1)},\bm{Y},\bm{Z}), \ldots,$ $(\bm{X}^{(M)},\bm{Y},\bm{Z})$ are exchangeable under $\mathcal{H}_0$, the above $p$-value \change{is valid, in the sense that it equals the significance level under the null, i.e.,}
\begin{eqnarray*}
\prob\left( p\le \alpha|\mathcal{H}_0 \right)= \alpha, \;\; \textrm{ for any } \; 0 < \alpha < 1. 
\end{eqnarray*}

In practice, however, $P_{X|Z}$ is rarely known. \cite{bellot2019conditional} proposed to approximate $P_{X|Z}$ using GANs. Specifically, they learned a generator $\mathbb{G}_X(\cdot,\cdot)$ from the observed data, then took $Z_i$ along with an independent noise variable as the input to obtain a sample $\widetilde{X}_i^{(m)}$, which minimizes the divergence between the distributions of $(X_i,Z_i)$ and $(\widetilde{X}_i^{(m)}, Z_i)$. They computed the $p$-value by replacing $\bm{X}^{(m)}$ with $\widetilde{\bm{X}}^{(m)}=(\widetilde{X}_1^{(m)},\ldots,\widetilde{X}_n^{(m)})^\top$. They called this test the \emph{generative conditional independence test} (GCIT). By Theorem 1 of \cite{bellot2019conditional}, the excess type-I error of this test is upper bounded as,
\begin{eqnarray}\label{eqn:excesspvalue}
\begin{split}
\prob\left( p\le \alpha|\mathcal{H}_0 \right) - \alpha & \le  \Mean \left\{ d_{\scriptsize{\hbox{TV}}}\left( \widetilde{P}_{\bm{X}|\bm{Z}} ,P_{\bm{X}|\bm{Z}} \right) \right\} \\
& = \Mean \left\{ \sup_A \left| \prob(\bm{X}\in A|\bm{Z})-\prob(\widetilde{\bm{X}}^{(m)}\in A|\bm{Z}) \right| \right\} \equiv D,
\end{split}
\end{eqnarray}
where $d_{\scriptsize{\hbox{TV}}}$ is the total variation norm between two probability distributions $P$ and $Q$ such that $d_{\scriptsize{\hbox{TV}}}(P,Q)=\sup_{A} |P(A)-Q(A)|$, the supremum is taken over all measurable sets $A$, and the expectations in (\ref{eqn:excesspvalue}) are taken with respect to $\bm{Z}$. 

By definition, the error term $D$ in (\ref{eqn:excesspvalue}) measures the quality of the conditional distribution approximation. \cite{bellot2019conditional} argued that this error term is negligible due to the capacity of deep neural networks in terms of estimating the conditional distribution. To the contrary, we find this approximation error is usually \emph{not} negligible, and consequently, it may inflate the type-I error and invalidate the test. We consider a simple example to further elaborate this.

\begin{example}
Suppose $X$ is one-dimensional, and follows a simple linear regression model, $X=Z^\top \beta_0+\varepsilon$, where the error $\varepsilon$ is independent of $Z$, and $\varepsilon\sim N(0,\sigma_0^2)$ for some $\sigma_0^2>0$. 
\end{example}

\noindent
Suppose we know a priori that the linear regression model holds. We thus estimate $\beta_0$ by ordinary least squares, and denote the resulting estimator by $\widehat{\beta}$. For simplicity, suppose $\sigma_0^2$ is known too. For this simple example, we have the following result regarding the approximation error $D$. 

\begin{proposition}\label{prop1}
Suppose the linear regression model holds, the dimension of $Z$ is much smaller than the sample size $n$, and the derived distribution $\widetilde{P}_{\bm{X}|\bm{Z}}$ is $\textrm{Normal}(\bm{Z} \widehat{\beta}, \sigma_0^2  I_n)$, where $I_n$ is the $n\times n$ identity matrix. Then $D$ does not decay to zero. 
\end{proposition}

To facilitate the understanding of the convergence behavior of $D$, we sketch a few lines of the proof of Proposition \ref{prop1}. The complete proof is given in the Appendix. Let $\widetilde{P}_{X|Z=Z_i}$ denote the conditional distribution of $\widetilde{X}^{(m)}_i$ given $Z_i$, which is $\textrm{Normal}(Z_i^\top \widehat{\beta},\sigma_0^2)$ in this example. If $D = o(1)$, then,
\begin{eqnarray}\label{eqn:tv}
\widetilde{D} \equiv n^{1/2} \sqrt{\Mean \left\{ d_{\scriptsize{\hbox{TV}}}^2 \left( \widetilde{P}_{X|Z=Z_i} ,P_{X|Z=Z_i} \right) \right\}} = o(1).
\end{eqnarray}  
In other words, \change{in order to control the type-I error}, GCIT requires the total variation distance measure in (\ref{eqn:tv}) to converge at a faster rate than $n^{-1/2}$. However, this rate cannot be achieved in general. In our Example 1, we have $\widetilde{D} \ge c$ for some constant $c>0$. Consequently, $D$ in (\ref{eqn:excesspvalue}) is not $o(1)$.  Proposition \ref{prop1} shows that, even if we know a priori that the linear model holds, $D$ does not decay to zero as $n$ tends to infinity. In practice, we do not have such prior model information. Then it would be even more difficult to estimate the conditional distribution $P_{X|Z}$. Therefore, using GANs to approximate $P_{X|Z}$ does not guarantee a negligible approximation error.

\subsection{Regression-based tests}

The family of regression-based tests is built upon the generalized covariance measure, 
\begin{eqnarray*}
\textrm{GCM}(X,Y) = \frac{1}{n} \sum_{i=1}^n \left\{ X_i-\widehat{\Mean} (X_i | Z_i) \right\} \left\{ Y_i-\widehat{\Mean} (Y_i | Z_i) \right\}, 
\end{eqnarray*}
where $\widehat{\Mean} (X|Z)$ and $\widehat{\Mean} (Y|Z)$ are the estimated condition means $\Mean (X|Z)$ and $\Mean (Y|Z)$, respectively, obtained by some supervised learner. When the prediction errors of $\widehat{\Mean} (X|Z)$ and $\widehat{\Mean} (Y|Z)$ satisfy certain convergence rates, \cite{shah2018hardness} proved that GCM is asymptotically normal under $\mathcal{H}_0$, in which the asymptotic mean is zero, and the standard deviation can be consistently estimated by some standard error estimator, denoted by $\widehat{s}(\textrm{GCM})$. Therefore, at level $\alpha$, we reject $\mathcal{H}_0$, if $|\textrm{GCM}| / \widehat{s}(\textrm{GCM})$ exceeds the upper $\alpha/2$th quantile of a standard normal distribution.

Such a test \change{can control the type-I error}. Nevertheless, it may not have sufficient power to detect $\mathcal{H}_1$. Consider the asymptotic mean of GCM, which is $\hbox{GCM}^*(X,Y) = \Mean \{X-\Mean(X|Z)\}\{Y-\Mean (Y|Z)\}$. The regression-based tests require $|\hbox{GCM}^*|$ to be nonzero under $\mathcal{H}_1$ to have power. However, it may be difficult to satisfy such a requirement. We again consider a simple example.

\begin{example} 
Suppose $X^*$, $Y$ and $Z$ are independent random variables. Besides, $X^*$ has mean zero, and $X=X^* g(Y)$ for some function $g$. 
\end{example}

\noindent
For this example, we have $\Mean(X|Z)=\Mean(X)$, since both $X^*$ and $Y$ are independent of $Z$, and so is $X$. Besides, $\Mean(X) = \Mean(X^*) \Mean\{g(Y)\} = 0$, since $X^*$ is independent of $Y$ and $\Mean(X^*)=0$. Thus $\textrm{GCM}^*(X, Y) = \Mean\{X-\Mean(X)\}\{Y-\Mean(Y|Z)\} = 0$ for any function $g$. On the other hand, $X$ and $Y$ are conditionally dependent given $Z$, as long as $g$ is not a constant function. Therefore, for this example, the regression-based tests would fail to discriminate between $\mathcal{H}_0$ and $\mathcal{H}_1$.

\subsection{MMD-based tests}

The family of MMD-based tests involves the maximum mean discrepancy as a measure of independence. For any two probability measures $P$, $Q$ and a function space $\mathbb{F}$, define
\begin{eqnarray*}
\textrm{MMD}(P,Q|\mathbb{F}) = \textrm{sup}_{f\in\mathbb{F}} \left\{ \Mean f(W_1) - \Mean f(W_2) \right\}, \;\; \textrm{ where } \; W_1\sim P, \; W_2\sim Q.
\end{eqnarray*}
Let $\mathbb{H}_1$, $\mathbb{H}_2$ denote some function spaces of $X$ and $Y$. Define 
\begin{eqnarray*}
\phi_{XY} = \hbox{MMD}(P_{XY},Q_{XY} \; | \; \mathbb{H}_1\otimes\mathbb{H}_2), 
\end{eqnarray*}
where $\otimes$ is the tensor product, $P_{XY}$ is the joint distribution of $(X,Y)$ whose definition does not rely on $Z$, and $Q_{XY}$ is the conditionally independent distribution with the same $X$ and $Y$ margins as $P_{XY}$. \change{Let $X'$ and $Y'$ be independent copies of $X$ and $Y$, such that they are conditionally independent given $Z$. Then $Q_{XY}$ corresponds to the joint distribution of $(X',Y')$. Note that, to generate $(X',Y')$, we need to first sample $Z$ according to $P_Z$, then generate $X'$ and $Y'$ that follow $P_{X|Z}$ and $P_{Y|Z}$, respectively. As such, $Q_{XY}$ depends on $Z$, and $\phi_{XY}$ depends on $Z$ through $Q_{XY}$.} Furthermore, since $\Mean \{h_1(X')h_2(Y')\} = \Mean [\Mean \{h_1(X')|Z\}\Mean \{h_2(Y')|Z\}]$, we have, 
\vspace{-0.05in}
\begin{align*}
\phi_{XY} = & \sup_{h_1\in\mathbb{H}_1,h_2\in\mathbb{H}_2} \left[ \Mean \{h_1(X)h_2(Y)\} - \Mean\{h_1(X')h_2(Y')\} \right] \\
= & \sup_{h_1\in\mathbb{H}_1,h_2\in\mathbb{H}_2} \Big( \Mean \{h_1(X)h_2(Y)\} - \Mean [\Mean \{h_1(X)|Z\}\Mean \{h_2(Y)|Z\}] \Big) \\
= & \sup_{h_1\in\mathbb{H}_1,h_2\in\mathbb{H}_2} \Big( \Mean \{h_1(X)h_2(Y)\} - \Mean [h_1(X) \Mean \{h_2(Y)|Z\}]-\Mean [\{h_1(X)|Z\}h_2(Y)] \\
& \quad\quad\quad\quad\quad\;\; + \Mean [\Mean \{h_1(X)|Z\}\Mean \{h_2(Y)|Z\}] \Big) \\
= & \sup_{h_1\in\mathbb{H}_1,h_2\in\mathbb{H}_2} \Mean \Big[h_1(X)-E\{h_1(X)|Z\}\Big]\Big[h_2(Y)-\Mean \{h_2(Y)|Z\}\Big].
\end{align*}
As such, $\phi_{XY}$ measures the average conditional association between $X$ and $Y$ given $Z$. Under $\mathcal{H}_0$, it equals zero, and hence an estimator of this measure can be used as a test statistic for $\mathcal{H}_0$. Moreover, if $\mathbb{H}_1$ and $\mathbb{H}_2$ are reproducing kernel Hilbert spaces (RKHSs), then $\phi_{XY}$ has a closed form expression in terms of the reproducing kernels of the RKHS \citep{doran2014permutation,gretton2012kernel}, which makes the tests based on an estimator of $\phi_{XY}$ easier to evaluate. 

\change{A notable example of this family is the kernel MMD-based test (KCIT) of \citet{zhang2011kernel}. We next further discuss this test. To control the type-I error asymptotically, KCIT requires the dimension $d_Z$ of $Z$ to be fixed \citep[Proposition 5]{zhang2011kernel}, since it uses the continuous mapping theorem to derive the limiting distribution of its test statistic. However, the continuous mapping theorem may not hold when $d_Z$ diverges with $n$. In addition, KCIT requires the $\ell_1$ distance between the covariance operator and its empirical estimator to decay to zero. It remains unknown whether such an assertion holds as $d_Z$ diverges. By contrast, the test we develop allows $d_Z$ to diverge while maintaining the asymptotic control of the type-I error. This implies that our test is expected to have a better size control than KCIT when $d_Z$ is large. We later further verify this through numerical simulations. Moreover, the maximization of KCIT is done over unit balls in an RKHS, while our proposed test can deal with much more general function spaces such as those generated by neural networks. Consequently, the power of our test can be tailored to more general alternatives than KCIT. For instance, it is known that deep neural networks learn certain non-smooth functions at a faster rate than kernel methods \citep{imaizumi2019deep}. This implies that our test is expected to have a better power than KCIT under certain types of alternatives.}

\section{A new double GANs-based testing procedure}
\label{sec:doubleGAN}

We propose a double GANs-based testing procedure for the conditional independence testing problem (\ref{eqn:test}). Conceptually, our test integrates three families of tests that are based on conditional randomization, regression, and MMD. Meanwhile, our new test overcomes the limitations of the existing ones. Unlike the GCIT of \cite{bellot2019conditional} that only learned the conditional distribution of $X$ given $Z$, we learn two generators $\mathbb{G}_X$ and $\mathbb{G}_Y$ to approximate the conditional distributions of both $X$ given $Z$ and $Y$ given $Z$. We then integrate the two generators in an appropriate way to construct a doubly-robust test statistic. To ensure the theoretical properties of this test, we only require the root mean squared total variation norm to converge at a rate of $n^{-\kappa}$ for some $\kappa>1/4$. Such a requirement is much weaker and practically more feasible than the condition in (\ref{eqn:tv}). 

Moreover, to improve the power of the test, we consider a collection of the generalized covariance measures, $\{\hbox{GCM}(h_1(X),h_2(Y)):h_1,h_2\}$, for multiple combinations of transformation functions $h_1(X)$ and $h_2(Y)$. We then take the maximum of all these GCMs as our test statistic. This essentially yields a type of maximum mean discrepancy measure $\phi_{XY}$. To see why this statistic can enhance the power, we quickly revisit Example 2. When $g$ is not a constant function, there exists some nonlinear function $h_1$ such that $h_1^*(Y)=\Mean \{h_1(X)|Y\}$ is not a constant function of $Y$. Set $h_2=h_1^*$. We then have $\textrm{GCM}^*=\Mean[h_1\{X^*g(Y)\}\{Y-\Mean(Y)\}] = \Var\{h_1^*(Y)\}>0$, which enables us to discriminate $\mathcal{H}_1$ from $\mathcal{H}_0$. 

We note that the maximum of GCMs yields MMD. Instead of using kernels, we have chosen GANs, because they have been shown to give good approximations of complex distributions \citep{imaizumi2019deep}. This allows the transformation functions $h_1$ and $h_2$ to be arbitrary function spaces. We set these function spaces to the class of neural networks in our implementation. In contrast, kernel based measures such as KCIT are limited to vector spaces of functions, which can be problematic for a high-dimensional conditioning variable \citep{doran2014permutation}.

We also remark that, even though our proposal is built upon the existing CI tests, our test is far from a simple extension. The major challenge lies in how to properly utilize the GAN estimators for the purpose of high-dimensional conditional independence testing. Despite the fact that GANs are capable of approximating complex high-dimensional probability distributions, the GAN estimators have non-negligible bias that decays slower than the parametric root-$n$ rate. Naively plugging the GAN estimators in the test statistic can invalidate the subsequent inference. 

\begin{figure}[t!]
\centering
\includegraphics[width=14.5cm]{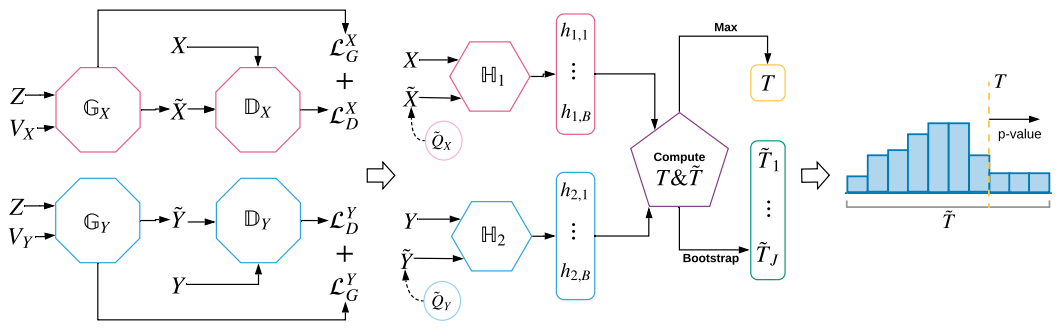}
\caption{Illustration of the conditional independence test with double GANs.}
\label{fig:overview}
\end{figure}

We give a graphical overview of our proposed testing procedure in Figure \ref{fig:overview}. We first employ double GANs to compute the test statistic that is the maximum of the GCMs over multiple transform functions. We then employ multiplier bootstrap to compute the corresponding $p$-value. We next detail the main components of our testing procedure. 

\subsection{Test statistic}

We begin with two function spaces, $\mathbb{H}_1 = \left\{ h_{1,\theta_1} : \theta_1 \in \mathbb{R}^{d_1} \right\}$ and $\mathbb{H}_2 = \left\{ h_{2,\theta_2} :\theta_2 \in \mathbb{R}^{d_2} \right\}$, indexed by some parameters $\theta_1$ and $\theta_2$, respectively.  In our implementation, we set $\mathbb{H}_1$ and $\mathbb{H}_2$ to the classes of neural networks with a single-hidden layer, finitely many hidden nodes, and the sigmoid activation function. However, a broad range of other function spaces may be considered, as appropriate for the application at hand. We next randomly generate $B$ functions, $h_{1,1}, \ldots, h_{1,B} \in \mathbb{H}_1$, $h_{2,1}, \ldots, h_{2,B} \in \mathbb{H}_2$, where we independently generate i.i.d.\ multivariate normal variables $\theta_{1,1}, \ldots, \theta_{1,B} \sim N(0, 2 I_{d_1}/d_1)$, and $\theta_{2,1}, \ldots, \theta_{2,B} \sim N(0, 2 I_{d_2}/d_2)$. We then set $h_{1,b}=h_{1,\theta_{1,b}}$, and $h_{2,b}=h_{2,\theta_{2,b}}$, $b \in [B] =\{1, \ldots, B\}$. Consider the following maximum-type test statistic,
\begin{eqnarray*}\label{eqn:teststat1}
T = \max_{b_1,b_2 \in [B]} \widehat{\sigma}_{b_1,b_2}^{-1} \left|\frac{1}{n}\sum_{i=1}^n \left[ h_{1,b_1}(X_i) -\widehat{\Mean }\{h_{1,b_1}(X_i)|Z_i\} \right] \left[ h_{2,b_2}(Y_i) -\widehat{\Mean }\{h_{2,b_2}(Y_i)|Z_i\} \right]\right|,
\end{eqnarray*}
where $\widehat{\sigma}_{b_1,b_2}^2$ is the sampling variance estimator,
\begin{eqnarray*}
\widehat{\sigma}_{b_1,b_2}^2=\frac{1}{n-1} \sum_{i=1}^n \bigg( \left[ h_{1,b_1}(X_i) -\widehat{\Mean }\{h_{1,b_1}(X_i)|Z_i\} \right] \left[ h_{2,b_2}(Y_i) -\widehat{\Mean }\{h_{2,b_2}(Y_i)|Z_i\} \right]\\- \frac{1}{n}\sum_{i=1}^n \left[ h_{1,b_1}(X_i) -\widehat{\Mean }\{h_{1,b_1}(X_i)|Z_i\} \right] \left[ h_{2,b_2}(Y_i) -\widehat{\Mean }\{h_{2,b_2}(Y_i)|Z_i\} \right] \bigg)^2.
\end{eqnarray*}
To compute $T$, we need to estimate the conditional means, $\Mean \{h_{1,b_1}(X)|Z\}$ and $\Mean \{h_{2,b_2}(Y)|Z\}$, which can be done by applying some supervised learning methods. However, this needs to be performed for all $b_1,b_2 \in [B]$. In theory, $B$ should diverge to infinity to guarantee the power property of the test. As such, this approach is computationally very expensive. Instead, we propose to implement this step based on the generators $\mathbb{G}_X$ and $\mathbb{G}_Y$ estimated using GANs, which is much more efficient computationally.  

Specifically, we first randomly generate i.i.d.\ samples $\{ v_{i,X}^{(m)}\}_{m=1}^{M}$, $\{v_{i,Y}^{(m)} \}_{m=1}^{M}$ from multivariate normal distribution, for $i = 1, \ldots, n$. We then feed $Z_i$ and $v_{i,X}^{(m)}$ into GANs to obtain the pseudo samples $\widetilde{X}_i^{(m)} = \mathbb{G}_X( Z_i,v_{i,X}^{(m)})$, and feed $Z_i$ and $v_{i,Y}^{(m)}$ to obtain $\widetilde{Y}_i^{(m)}=\mathbb{G}_Y( Z_i,v_{i,Y}^{(m)})$, for $i = 1, \ldots, n, m=1,\ldots,M$. These pseudo samples approximate the conditional distribution of $X_i$ and $Y_i$ given $Z_i$, respectively. We then compute 
\begin{eqnarray*}
\widehat{\Mean} \{h_{1,b_1}(\widetilde{X}_i)|Z_i\} = \frac{1}{M} \sum_{m=1}^{M} h_{1,b_1}( \widetilde{X}_i^{(m)}), \quad \widehat{\Mean} \{h_{2,b_2}(Y_i)|Z_i\} = \frac{1}{M}\sum_{m=1}^{M} h_{2,b_2}(\widetilde{Y}_i^{(m)}),
\end{eqnarray*}
for $b_1,b_2 = 1, \ldots, B$. Plugging the estimated means into $T$ produces the sample test statistic, 
\begin{eqnarray}\label{eqn:hatT}
& & \widehat{T} = \max_{b_1,b_2} \left| n^{-1/2} \sum_{i=1}^n \psi_{b_1,b_2,i} \right|, \quad \textrm{ where }  \\\nonumber
& & \psi_{b_1,b_2,i} = \widehat{\sigma}_{b_1,b_2}^{-1} \left\{h_{1,b_1}(X_i)-\frac{1}{M}\sum_{m=1}^{M} h_{1,b_1}\left( \widetilde{X}_i^{(m)} \right) \right\}  \left\{h_{2,b_2}(Y_i)-\frac{1}{M}\sum_{m=1}^{M} h_{2,b_2}\left( \widetilde{Y}_i^{(m)} \right) \right\}. 
\end{eqnarray}

To help reduce the type-I error, we further employ a data splitting and cross-fitting strategy, which has been commonly used in statistical inferences in recent years \citep{romano2019}. That is, we use different subsets of data samples to learn GANs and to construct the test statistic. We begin by dividing the data into $L$ folds of equal size. We use $\mathcal{I}^{(\ell)}$ to denote the set of indices of subsamples in the $\ell$th fold, and $\mathcal{I}^{(-\ell)}$ its complement. We next learn two generators $\mathbb{G}_X^{(\ell)}$ and $\mathbb{G}_Y^{(\ell)}$, based on $\{(X_i,Z_i)\}_{i\in \mathcal{I}^{(-\ell)}}$ and $\{(Y_i,Z_i)\}_{i\in \mathcal{I}^{(-\ell)}}$, to approximate the conditional distributions of $X|Z$ and $Y|Z$, for $\ell=1,\cdots,L$. Finally, for each $\ell$ and $i\in \mathcal{I}^{(\ell)}$, we generate the pseudo samples $\widetilde{X}_i^{(m)}$ and $\widetilde{Y}_i^{(m)}$ using $\mathbb{G}_X^{(\ell)}$ and $\mathbb{G}_Y^{(\ell)}$, and construct $\widehat{T}$ as in \eqref{eqn:hatT}. In this way, $\widetilde{X}_i^{(m)}$ and $\widetilde{Y}_i^{(m)}$ are conditionally independent of the observations in $\mathcal{I}^{(\ell)}$ given $Z_i$. Such a cross-fitting strategy allows us to derive the asymptotic properties of the test under minimal conditions on the generators. 

We summarize our procedure of computing the test statistic in Algorithm \ref{alg1}. 

\begin{algorithm}[t!]
\begin{algorithmic}
\item[\textbf{Input:}] The number of transformation functions $B$, the number of pseudo samples $M$, and the number of data splits $L$.
		
\item[\textbf{Step 1:}] Divide $\{1,\ldots,n\}$ into $L$ folds $\mathcal{I}^{(1)},\ldots,\mathcal{I}^{(L)}$. Denote $\mathcal{I}^{(-\ell)}=\{1,\ldots,n\} \backslash \mathcal{I}^{(\ell)}$. 
		
\item[\textbf{Step 2:}] For $\ell=1,\ldots,L$, train two generators $\mathbb{G}_X^{(\ell)}$ and $\mathbb{G}_Y^{(\ell)}$ based on $\{(X_i,Z_i)\}_{i\in \mathcal{I}^{(-\ell)}}$ and $\{(Y_i,Z_i)\}_{i\in \mathcal{I}^{(-\ell)}}$, to approximate the conditional distributions of $X|Z$ and $Y|Z$.
		
\item[\textbf{Step 3:}] For $\ell=1,\ldots,L$ and $i\in \mathcal{I}_{\ell}$, generate i.i.d.\ random noises $\left\{ v_{i,X}^{(m)} \right\}_{m=1}^M$, $\left\{ v_{i,Y}^{(m)} \right\}_{m=1}^M$. Set $\widetilde{X}_i^{(m)} = \mathbb{G}_X^{(\ell)}\left( Z_i,v_{i,X}^{(m)} \right)$, and $\widetilde{Y}_i^{(m)} = \mathbb{G}_Y^{(\ell)}\left( Z_i,v_{i,Y}^{(m)} \right)$, $m = 1, \ldots, M$.
		
\item[\textbf{Step 4:}] Randomly generate $h_{1,1},\ldots,h_{1,B}\in \mathbb{H}_1$ and $h_{2,1},\ldots,h_{2,B}\in \mathbb{H}_2$. 
		
\item[\textbf{Step 5:}] Compute the test statistic $\widehat{T}$. 
\end{algorithmic}
\caption{Algorithm for computing the test statistic.}
\label{alg1}
\end{algorithm}

\subsection{Approximation of conditional distribution via GANs}
\label{secGAN}

There are numerous GANs methods available for learning high-dimensional distributions. We adopt the proposal of \cite{genevay2017learning} to learn the conditional distributions $P_{X|Z}$ and $P_{Y|Z}$ in our setting thanks to its competitive performance. Recall that $\widetilde{P}_{X|Z}$ is the  distribution of pseudo outcome generated by the generator $\mathbb{G}_X$ given $Z$. We consider estimating $P_{X|Z}$ by optimizing 
\begin{eqnarray*}
\min_{\mathbb{G}_X} \max_{c} \mathcal{\widetilde{D}}_{c, \epsilon}(P_{X|Z}, \widetilde{P}_{X|Z}).
\end{eqnarray*}
Here $\mathcal{\widetilde{D}}_{c, \epsilon}$ denotes the Sinkhorn loss function between two probability measures with respect to some cost function $c$ and some regularization parameter $\epsilon>0$,
\begin{eqnarray*}
\mathcal{\widetilde{D}}_{c, \epsilon}(\mu, \nu) & = & 2\mathcal{D}_{c, \epsilon}(\mu, \nu) - \mathcal{D}_{c, \epsilon}(\mu, \mu) - \mathcal{D}_{c, \epsilon}(\nu, \nu), \\
\mathcal{D}_{c, \epsilon}(\mu, \nu) & = & \inf_{\pi \in \Pi(\mu, \nu)} \int_{x,y} \big\{ c(x, y)-\epsilon H(\pi|\mu  \otimes \nu) \big\} \pi(dx,dy),
\end{eqnarray*}
where $\Pi(\mu, \nu)$ is a set containing all probability measures $\pi$ whose marginal distributions correspond to $\mu$ and $\nu$, $H$ is the Kullback-Leibler divergence, and $\mu \otimes \nu$ is the product measure of $\mu$ and $\nu$. When $\epsilon = 0$, $\mathcal{D}_{c, 0}(\mu, \nu)$ measures the optimal transport of $\mu$ into $\nu$ with respect to the cost function $c(\cdot, \cdot)$ \citep{cuturi2013sinkhorn}.  When $\epsilon \neq 0$, an entropic regularization is added to this optimal transport. As such, the objective function $\mathcal{D}_{c, \epsilon}$ is a regularized optimal transport metric, and the regularization is to facilitate the computation, so that $\mathcal{D}_{c, \epsilon}$ can be efficiently evaluated.

Intuitively, the closer the two probability measures, the smaller the Sinkhorn loss. As such, maximizing the loss with respect to the cost function  learns a discriminator that can better discriminate the samples generated between $P_{X|Z}$ and $\widetilde{P}_{X|Z}$. On the other hand, minimizing the maximum cost with respect to the generator $\mathbb{G}_X$ makes it closer to the true distribution $P_{X|Z}$. This yields the minimax formulation $\min_{\mathbb{G}_X} \max_{c} \mathcal{\widetilde{D}}_{c, \epsilon}(P_{X|Z}, \widetilde{P}_{X|Z})$ that we target. In practice, we approximate the cost and the generator based on neural networks. Integrations in the objective function $\mathcal{\widetilde{D}}_{c, \epsilon}(P_{X|Z}, \widetilde{P}_{X|Z})$ are approximated by sample averages. The conditional distribution of $P_{Y|Z}$ is estimated similarly.

\subsection{Bootstrap for the $p$-value}

Next, we propose a multiplier bootstrap method to approximate the distribution of $\widehat{T}$ under $\mathcal{H}_0$ and compute the corresponding $p$-value. Let $\psi_{b_1,b_2}=n^{-1}  \sum_{i=1}^n \psi_{b_1,b_2,i}$. The key observation is that $\{\psi_{b_1,b_2}\}_{b_1,b_2=1}^{B}$ are asymptotically multivariate normal with zero mean under $\mathcal{H}_0$; see the proof of Theorem \ref{thm3} for details. Consequently, $\widehat{T} = \max_{b_1,b_2} |n^{-1/2} \sum_{i=1}^n \psi_{b_1,b_2,i}|$ is to converge to a maximum of normal variables in absolute values. 

To approximate this limiting distribution, we first estimate the covariance matrix of a $B^2$-dimensional vector formed by $\{n^{-1/2}\psi_{b_1,b_2}\}_{b_1,b_2=1}^{B}$ using the sample covariance matrix $\widehat{\Sigma}$, whose $\{b_1+B(b_2-1),b_3+B(b_4-1)\}$th entry is given by 
\begin{eqnarray*}
	\frac{1}{n} \sum_{i=1}^n (\psi_{b_1,b_2,i}-\psi_{b_1,b_2})(\psi_{b_3,b_4,i}-\psi_{b_3,b_4}), \quad b_1,b_2,b_3,b_4 = 1, \ldots, B.
\end{eqnarray*}
We then generate i.i.d.\ random vectors with the covariance matrix equal to $\widehat{\Sigma}$. This can be achieved by generating i.i.d.\ standard normal variables $\{W_{i,j}\}_{i,j}$ for $1\le i\cdots\le n$ and $j=1,\cdots,J$, then compute $B^2$-dimensional normal random vectors $\bm{W}_j$ whose $\{b_1+B(b_2-1)\}$th entry is given by $n^{-1/2} \sum_{i=1}^n (\psi_{b_1,b_2,i}-\psi_{b_1,b_2}) W_{i,j}$ for $j=1,\cdots,J$.  We next compute $\widetilde{T}_j=\| \bm{W}_j\|_{\infty}$, for $j=1,\ldots,J$, where $\|\cdot\|_{\infty}$ is the maximum element of a vector in absolute value, and $J$ is the number of bootstrap samples. Finally, we use these maximum absolute values to approximate the distribution of $\widehat{T}$ under the null hypothesis. This yields the $p$-value, $p=J^{-1}\sum_{j=1}^J \mathbb{I}(\widehat{T} \ge \widetilde{T}_j)$. We summarize this bootstrap procedure in Algorithm \ref{alg2}. 

\begin{algorithm}[b!]
\begin{algorithmic}
\item[\textbf{Input:}] The number of bootstrap samples $J$, and $\{\psi_{b_1,b_2,i}\}_{b_1,b_2=1,i=1}^{B,n}$. 
		
\item[\textbf{Step 1:}] Generate i.i.d.\ standard normal variables $W_{i,j}$ for $i=1,\cdots,n$, $j=1,\ldots,J$.
		
\item[\textbf{Step 2:}] Compute $B^2$-dimensional normal random vectors $\bm{W}_j$ whose $\{b_1+B(b_2-1)\}$th entry is given by $n^{-1/2} \sum_{i=1}^n (\psi_{b_1,b_2,i}-\psi_{b_1,b_2}) W_{i,j}$ and set $\widetilde{T}_j=\| \bm{W}_j\|_{\infty}$ for $j=1,\cdots,J$.
		
\item[\textbf{Step 3:}] Compute the $p$-value, $p=J^{-1}\sum_{j=1}^J \mathbb{I}(\widehat{T} \ge \widetilde{T}_j)$.
\end{algorithmic}	
\caption{Algorithm for computing the $p$-value.}
\label{alg2}
\end{algorithm}

\section{Asymptotic theory}
\label{sec:theory}

To derive the theoretical properties of the test statistic $\widehat{T}$, we first introduce the concept of the ``oracle" test statistic $T^*$. If $P_{X|Z}$ and $P_{Y|Z}$ were known a priori, then one can draw $\{X_i^{(m)}\}_{m}$ and $\{Y_i^{(m)}\}_{m}$ from $P_{X|Z=Z_i}$ and $P_{Y|Z=Z_i}$ directly, and can compute the test statistic by replacing $\{\widetilde{X}_{i}^{(m)}\}_m$ and $\{\widetilde{Y}_i^{(m)}\}_m$ with $\{X_i^{(m)}\}_m$ and $\{Y_i^{(m)}\}_m$. We call the resulting $T^*$ an ``oracle" test statistic. We next establish the double-robustness property of $\widehat{T}$, which helps explain why our test can relax the requirement in (\ref{eqn:tv}). Roughly  speaking, the double-robustness means that $\widehat{T}$ is asymptotically equivalent to $T^*$ when either the conditional distribution of $X | Z$, or that of $Y | Z$, is well approximated by GANs. It guarantees that $\widehat{T}$ converges to $T^*$ at a faster rate than the estimated conditional distribution. In contrast, the convergence rate of the GCIT test statistic is the same as the rate of the estimated conditional distribution. For this reason, our procedure only requires a weaker condition. 

\begin{theorem}[Double-robustness]\label{thm1}
Suppose $M$ is proportional to $n$, and $B=O(n^c)$ for some constant $c>0$. Suppose $\min_{h_1\in \mathbb{H}_1,h_2\in \mathbb{H}_2} \Var  [\{h_1(X) -\Mean \{h_1(X)|Z\} \}\{h_2(Y) -\Mean \{h_2(Y)|Z\} \}]\ge c^*$ for some constant $c^*>0$. Then, $\widehat{T}-T^*=o_p(1)$, when 
\begin{eqnarray*}
\Mean \left[ d_{\scriptsize{\hbox{TV}}}^2\left\{ \widetilde{Q}^{(\ell)}_X(\cdot|Z) ,Q_X(\cdot|Z) \right\} \right]  = o(\log^{-1} n), \, \textrm{ or } \,
\Mean \left[ d_{\scriptsize{\hbox{TV}}}^2\left\{ \widetilde{Q}^{(\ell)}_Y(\cdot|Z) ,Q_Y(\cdot|Z) \right\} \right] = o(\log^{-1} n).
\end{eqnarray*} 
\end{theorem}

\noindent
We note that the conditions on $M$ and $B$ are mild, as these are user-specified parameters. As we have mentioned, when both total variation distances converge to zero, the test statistic $T$ converges at a faster rate than those total variation distances. Therefore, we can greatly relax the condition in (\ref{eqn:tv}), and replace it with, 
\begin{eqnarray}\label{eqn:tv2}
\left[ \Mean\left\{ d_{\scriptsize{\hbox{TV}}}^2\left( \widetilde{P}_{X|Z}^{(\ell)} ,P_{X|Z} \right) \right\} \right]^{1/2} = O(n^{-\change{\kappa_x}}), \; \textrm{and} \;
\left[ \Mean\left\{ d_{\scriptsize{\hbox{TV}}}^2\left( \widetilde{P}_{Y|Z}^{(\ell)} ,P_{Y|Z} \right) \right\} \right]^{1/2} = O(n^{-\change{\kappa_y}}), 
\end{eqnarray}
for some constants $0<\kappa_x,\kappa_y<1/2$ and any $\ell \in [L]$, where $\widetilde{P}_{X|Z}^{(\ell)}$ and $\widetilde{P}_{Y|Z}^{(\ell)}$ denote the conditional distributions approximated via GANs trained on the $\ell$-th subset of data samples. The next theorem summarizes this discussion.

\begin{theorem}\label{thm2}
Suppose the conditions in Theorem \ref{thm1}. Furthermore, suppose \eqref{eqn:tv2} holds. Then, $\widehat{T}-T^* = O_p\left( n^{-(\change{\kappa_x+\kappa_y})}\log n \right)$. 
\end{theorem}

Since $\kappa_x,\kappa_y>0$, the convergence rate of $(\widehat{T}-T^*)$ is faster than that in (\ref{eqn:tv2}). To ensure $\sqrt{n}(T-T^*)=o_p(1)$, it suffices to require \change{$\kappa_x+\kappa_y>1/2$}. In contrast to (\ref{eqn:tv}), this rate is achievable. We consider two examples in \cite{berrett2019conditional} to illustrate this, while the condition holds in a much wider range of settings. 

\begin{example}[Parametric setting] 
\change{Suppose the parametric forms of $Q_X$ and $Q_Y$ are correctly specified. Then under certain regularity conditions, the requirement $\kappa_x+\kappa_y>1/2$ holds if $k_x=O(n^{t_x})$ and $k_y=O(n^{t_y})$ for some $t_x+t_y<1/2$, where $k_x$ and $k_y$ are the dimensions of the parameters defining the parametric models for $Q_X$ and $Q_y$, respectively.} 
\end{example}

\begin{example}[Nonparametric setting with binary data]
\change{Suppose $X, Y$ are binary variables. Then the requirement $\kappa_x+\kappa_y>1/2$ holds if the mean squared prediction errors of the nonparametric estimators of the conditional means of $X$ and $Y$ given $Z$ are $O(n^{-t_x})$ and $O(n^{-t_y})$ for some $t_x$, $t_y$, such that $t_x+t_y>1/2$.}   
\end{example}

We briefly remark that, \change{there is no explicit specification on $d_Z$ in the statement of Theorem \ref{thm2}. It is implicitly imposed due to the requirement that $\kappa_x+\kappa_y>1/2$, and $d_Z$ is allowed to diverge with the sample size.} In addition, the condition \change{$\kappa_x+\kappa_y>1/2$} can be further relaxed to \change{$\kappa_1,\kappa_2>0$} using the theory of higher order influence functions \citep{robins2008higher,robins2017minimax,mukherjee2017semiparametric}. However, the resulting estimators would be considerably much more complicated, and thus we do not pursue those estimators.  
 
Next, we show that our proposed test can control the type-I error asymptotically.

\begin{theorem}\label{thm3}
Suppose the conditions in Theorem \ref{thm1} hold. Suppose (\ref{eqn:tv2}) holds for some \change{$\kappa_x$, $\kappa_y$ such that $\kappa_x+\kappa_y>1/2$}. Then, the $p$-value from Algorithm \ref{alg2} satisfies that $\prob(p\le \alpha|\mathcal{H}_0)= \alpha+o(1)$. 
\end{theorem}

Next, to derive the asymptotic power of the test, we introduce the pair of hypotheses based on the notion of weak conditional independence \citep{daudin1980partial}, 
\begin{eqnarray*}
	&&\mathcal{H}_0^*: \Mean [ \Cov\{ f(X),g(Y)|Z \} ] = 0, \;\; \textrm{ for any } f \in L_X^2, g \in L_Y^2 \quad \textrm{versus}  \\
	&&\mathcal{H}_1^*: \Mean [ \Cov\{ f(X),g(Y)|Z \} ] \neq 0, \;\; \textrm{ for some } f \in L_X^2, g \in L_Y^2,
\end{eqnarray*}
where $L_X^2$ and $L_Y^2$ denote the class of all squared integrable functions of $X$ and $Y$, respectively. We note that conditional independence implies weak conditional independence, i.e., $\mathcal{H}_0$ implies $\mathcal{H}_0^*$, and $\mathcal{H}_1^*$ implies $\mathcal{H}_1$. \change{We consider an example to further elaborate on the difference between weak CI and CI.

\begin{example}
Let $X,Y,Z$ be binary random variables with the distribution functions,
\begin{eqnarray*}
	\left(\begin{array}{cc}
		\prob(X=0,Y=0|Z=0) & \prob(X=0,Y=1|Z=0) \\
		\prob(X=1,Y=0|Z=0) & \prob(X=1,Y=1|Z=0)
	\end{array}
	\right)=\left(\begin{array}{cc}
		1/6 & 1/3 \\
		1/3 & 1/6
	\end{array}
	\right),\\
	\left(\begin{array}{cc}
		\prob(X=0,Y=0|Z=1) & \prob(X=0,Y=1|Z=1) \\
		\prob(X=1,Y=0|Z=1) & \prob(X=1,Y=1|Z=1)
	\end{array}
	\right)=\left(\begin{array}{cc}
		1/3 & 1/6 \\
		1/6 & 1/3
	\end{array}
	\right),
\end{eqnarray*}
and $Z$ takes the value $\{0,1\}$ with equal probability. We can show that, for any $x,y\in \{0,1\}$, 
\begin{eqnarray*}
\Mean \{\prob(X=x|Z) \prob(Y=y|Z)\}=\frac{1}{2}\times \frac{1}{2}=\frac{1}{4}, \\
\prob(X=x,Y=y)=\frac{1}{2} \Big\{ \prob(X=x,Y=y|Z=0)+\prob(X=x,Y=y|Z=1) \Big\}
\\ = \frac{1}{2}\times \left(\frac{1}{6}+\frac{1}{3}\right)=\frac{1}{4}.
\end{eqnarray*}
By definition, this implies that $X$ and $Y$ are weakly conditionally independent given $Z$, since
\begin{align*}
\Mean [\Cov\{f(X),g(Y)|Z\} ]=\sum_{x,y} f(x)g(y) \Big\{ & \prob(X=x,Y=y) \\ 
& - \Mean \big\{ \prob(X=x|Z) \prob(Y=y|Z) \big\} \Big\} =0.
\end{align*}
However, $\prob(X=0,Y=0|Z=0) \neq \prob(X=0|Z=0)\prob(Y=0|Z=0)$, since the former equals $1/6$, and the latter equals $1/4$. As such, $X$ and $Y$ are not conditionally independent given $Z$.
\end{example}
}

The next theorem shows that our proposed test is consistent against the alternatives in $\mathcal{H}_1^*$, but not against all alternatives in $\mathcal{H}_1$. 

\begin{theorem}\label{thm4}
Suppose the conditions in Theorem \ref{thm3} hold, $B=c_0n^c$ for some $c_0,c>0$, and $X$, $Y$ are bounded random variables. Then the $p$-value from Algorithm \ref{alg2} satisfies that $\prob(p\le \alpha|\mathcal{H}_1^*)\to 1$, as $n\to \infty$. 
\end{theorem}

Finally, we remark that our test is constructed based on $\phi_{XY}$. Meanwhile, we may consider another test based on $\phi_{XYZ} = \hbox{MMD}(P_{XYZ},Q_{XYZ}|\mathbb{H}_1\otimes\mathbb{H}_2\otimes\mathbb{H}_3)$, where $P_{XYZ}$ is the joint distribution of $(X,Y,Z)$, $Q_{XYZ}=P_{X|Z}P_{Y|Z}P_Z$, and $\mathbb{H}_3$ is a neural network class of functions of $Z$. This type of test is consistent against all alternatives in $\mathcal{H}_1$. However, in our numerical experiments, we find it  less powerful compared to our test. This agrees with the observation by \cite{li2019nonparametric} in that, even though the tests based on weak CI cannot fully characterize CI, they potentially enjoy an improved power.

\section{Numerical studies}
\label{sec:numeric}

We begin with a discussion of some implementation details. We then carry out simulations to study the empirical size and power of the proposed test, and compare with several alternative methods. We further illustrate with an application to a cancer genetics example.

\subsection{Implementation details}

For the number of functions $B$ in Algorithm \ref{thm1}, it represents a trade-off. By Theorem \ref{thm4}, $B$ should be as large as possible to guarantee a good power. In practice, the computation complexity increases as $B$ increases. Our numerical studies suggest that the value of $B$ between $30$ and $50$ achieves a good balance between the power and the computational cost, and we fix $B=30$. For the number of pseudo samples $M$, and the number of sample splittings $L$, we find the results are not overly sensitive to their choices, and thus we fix $M=100$ and $L=3$. \change{Besides, we set the number of bootstrap samples $J = 1000$.}

\begin{figure}[t!]
\begin{tabular}{cc}
\includegraphics[width=7cm]{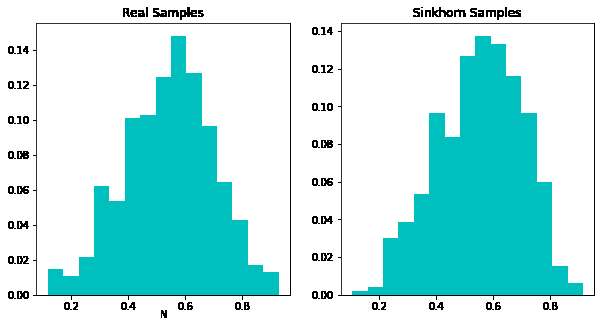}& 
\includegraphics[width=7cm]{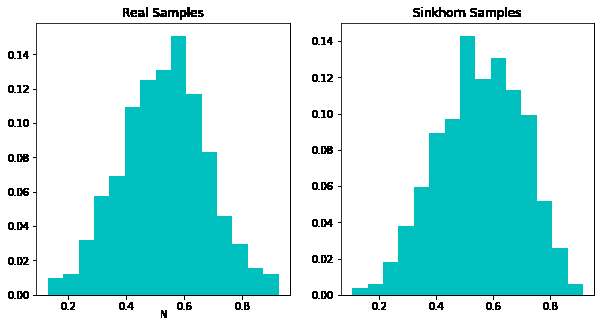}\\
(a) One random value of $a_g$ & (b) Another random value of $a_g$
\end{tabular}
\caption{Conditional histograms. GANs are trained using data generated from the simulation study in Section \ref{sec:simu}.}
\label{fig:ch}	
\end{figure}

For the GANs, we use a single-hidden layer neural network to approximate both the discriminator and the generator. The number of nodes in the hidden layer is set at $128$. The dimension of the input noise $v_{i,X}^{(m)}$ and $v_{i,Y}^{(m)}$ is set at $10$. These tuning parameters are chosen following the common practice in the GANs literature, and also by investigating the goodness-of-fit of the resulting generator, which can be done by comparing the conditional histogram of the generated samples to that of the true samples. In our experiments, we find such an approach yields GANs with satisfactory performances. More specifically, let $d_Z$ denote the dimension of $Z$, and $\widehat{\mu}_Z$ the sample average $n^{-1}\sum_i Z_i$. {Let $\widetilde{Y}_i=G_Y(Z_i, v_{i,Y})$ denote a simulated sample to approximate the distribution of $Y|Z=Z_i$ obtained by the generator $G_Y$. When $G_Y$ is accurate, we expect the conditional distribution of $\widetilde{Y}_i$ and $Y_i$ given $Z_i$ are similar. As such, for any $d_Z$-dimensional vector $a$, the histograms $\{\widetilde{Y}_i: a^\top (\widetilde{Z}_i-\widehat{\mu}_Z)>0\}$ and $\{Y_i: a^\top (Z_i-\widehat{\mu}_Z)>0\}$ should be similar}. We sample i.i.d.\ vectors $\{a_g\}_g$ from $\textrm{Normal}(0, I_{d_Z})$. For each $g$, we plot the histogram $\{Y_i: a_g^\top (Z_i-\widehat{\mu}_Z)>0\}$ and $\{\widetilde{Y}_i^{(m)}: a_g^\top (Z_i-\widehat{\mu}_Z)>0\}$. See Figures \ref{fig:ch} (a) and (b) for the conditional histograms with two choices of $a_g$. It is seen that the GANs fit the conditional density reasonably well. The fitted conditional distribution for $P_{X|Z}$ can be checked in a similar fashion.

\subsection{Simulations}
\label{sec:simu}

We generate the data following the post nonlinear noise model similarly as in \cite{zhang2011kernel, doran2014permutation, bellot2019conditional}, i.e., 
\begin{eqnarray*}
X = \sin(a_f^\top Z + \varepsilon_f), \quad \textrm{and} \quad Y = \cos(a_g^\top Z + b X+\varepsilon_g). 
\end{eqnarray*}
The entries of $a_f, a_g$ are randomly and uniformly sampled from $[0,1]$, then normalized to the unit $\ell_1$ norm. The noise variables $\varepsilon_f, \varepsilon_g$ are independently sampled from a normal distribution with mean zero and variance $0.25$. In this model, the parameter $b$ determines the degree of conditional dependence. When $b=0$, $\mathcal{H}_0$ holds, and otherwise $\mathcal{H}_1$ holds. The sample size is set at $n=1000$. 

We call our test DGCIT, short for double GANs-based conditional independence test. We compare it with the GCIT test of \cite{bellot2019conditional}, the regression-based test (RCIT) of \cite{shah2018hardness}, the kernel MMD-based test (KCIT) of \cite{zhang2011kernel}, and the classifier CI test (CCIT) of \cite{sen2017model}. 

\begin{figure}[t!]
\centering
\includegraphics[width=6cm, height=5cm]{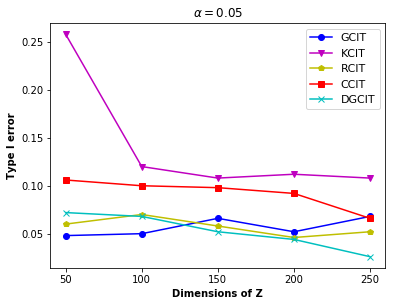}\hspace{-0.15cm}
\includegraphics[width=6cm, height=5cm]{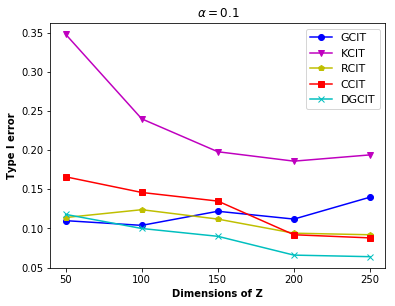}\hspace{-0.15cm}
\includegraphics[width=6cm, height=5cm]{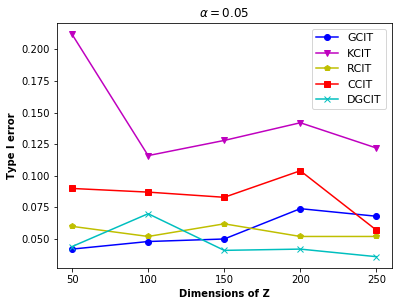}\hspace{-0.15cm}
\includegraphics[width=6cm, height=5cm]{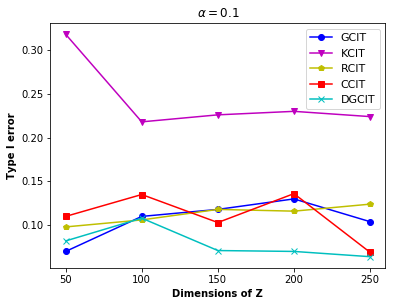}
\caption{The empirical type-I error rate of various tests under $\mathcal{H}_0$. Left panels: $\alpha=0.05$, right panels: $\alpha=0.1$. Top panels: $Z$ is normal, bottom panels: $Z$ is Laplacian.}
\label{fig-sim}
\end{figure}

We first study the empirical size when $b=0$. We vary the dimension of $Z$ as $d_Z = 50, 100, 150, 200, 250$, and consider two generation distributions. We first generate $Z$ from a standard normal distribution, then from a Laplace distribution. We set the significance level at $\alpha=0.05$ and $0.1$. Figure \ref{fig-sim} reports the empirical size of the tests aggregated over 500 data replications. We make the following observations. First, the type-I error rates of our test and RCIT are close to or below the nominal level in nearly all cases. Second, KCIT  fails in that its type-I error is considerably larger than the nominal level in all cases. \change{We suspect it is due to the high-dimensional setting where $d_Z \ge 50$. We have experimented with $d_Z = 5$, and found that KCIT is able to control the type-I error in that case. This is consistent with Proposition 5 of \cite{zhang2011kernel}, which suggests that KCIT should work in a low-dimensional setting.}  Third, GCIT and CCIT both have inflated type-I errors in some cases. Take GCIT as an example. When $Z$ is normal, $d_Z=250$ and $\alpha=0.1$, its empirical size is close to $0.15$. This is consistent with our discussion in Section \ref{secCRT}, since GCIT requires a rather strong condition to control the type-I error. 

\begin{figure}[t!]
\centering
\includegraphics[width=6cm, height=5cm]{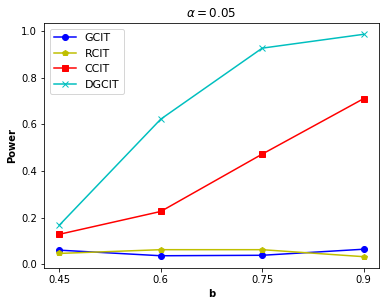}\hspace{-0.15cm}
\includegraphics[width=6cm, height=5cm]{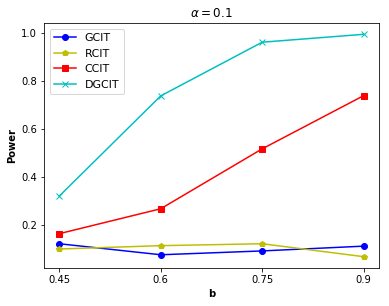}\hspace{-0.15cm}
\includegraphics[width=6cm, height=5cm]{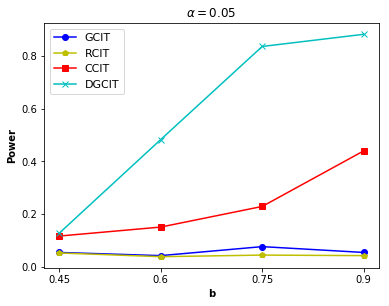}\hspace{-0.15cm}
\includegraphics[width=6cm, height=5cm]{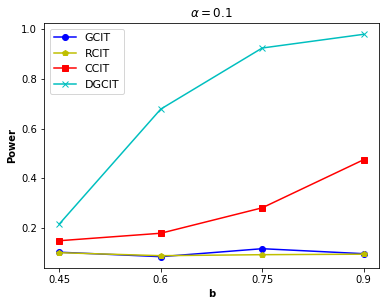}
\caption{The empirical power of various tests under $\mathcal{H}_1$. Left panels: $\alpha=0.05$, right panels: $\alpha=0.1$. Top panels: $d_Z=100$, bottom panels: $d_Z=200$.}
\label{fig-sim1}
\end{figure}

We then study the empirical power when $b>0$. We generate $Z$ from a standard normal distribution, with $d_Z = 100, 200$, and vary the value of $b = 0.3, 0.45, 0.6, 0.75, 0.9$ that controls the magnitude of the alternative. Figure \ref{fig-sim1} reports the empirical power of the tests over 500 data replications. We observe that our test is the most powerful, and the empirical power approaches 1 as $b$ increases to $0.9$, demonstrating the consistency of the test. Meanwhile, both GCIT and RCIT have no power in all cases. We do not report the power of KCIT, because as we have shown earlier, it cannot control the size, and thus its empirical power is not meaningful.  

Finally, we discuss the computation time. All experiments were run on a 16 N1 CPUs Google Cloud Computing platform. \change{The wall clock time for running the entire GCIT test for one data replication was about 2.5 minutes. In contrast, the running time for CCIT was about 2 minutes, for KCIT about 30 seconds, and for GCIT and RCIT  about 20 seconds.} 

\subsection{Anti-cancer drug data example}

We illustrate our proposed test with an anti-cancer drug dataset from the Cancer Cell Line Encyclopedia \citep{barretina2012cancer}. We concentrate on a subset, the CCLE data, that measures the treatment response of drug PLX4720. It is well known that the  patient's cancer treatment response to drug can be strongly influenced by alterations in the genome \citep{Garnett2012}. This data measures 1638 genetic mutations of $n=472$ cell lines, and the goal of our analysis is to determine which genetic mutation is significantly correlated with the drug response after conditioning on all other mutations. The same data was also analyzed in \cite{tansey2018holdout} and \cite{bellot2019conditional}. We adopt the same screening procedure as theirs to screen out irrelevant mutations, which leaves a total of 466 potential mutations for our conditional independence testing.

\begin{table}[b!]
\caption{The variable importance measures of the elastic net and random forest models, versus the $p$-values of the GCIT and DGCIT tests for the anti-cancer drug example.}\label{tab1}
\resizebox{\textwidth}{!}{
\begin{tabular}{ccccccccccc}\toprule
& BRAF.V600E &  BRAF.MC & HIP1 & FTL3  & CDC42BPA & THBS3 & DNMT1 & PRKD1 & PIP5K1A & MAP3K5 \\ \midrule
EN & 1 & 3 & 4 & 5 & 7 & 8 & 9 & 10  & 19 & 78 \\
RF & 1 & 2 & 3 & 14 & 8 & 34 & 28 & 18 & 7 & 9\\\midrule 
GCIT & $<$0.001 & $<$0.001 & 0.008 & 0.521 & 0.050 & 0.013 & 0.020 & 0.002 & 0.001 & $<$0.001 \\
DGCIT & 0 & 0 & 0 & 0 & 0 & 0 & 0 & 0 & 0 & 0.794\\ \bottomrule
\end{tabular}
}
\end{table}

The ground truth is unknown for this data. Instead, we compare with the variable importance measures obtained from fitting an elastic net (EN) model and a random forest (RF) model as reported in \cite{barretina2012cancer}. In addition, we compare with the GCIT test of \cite{bellot2019conditional}. Table \ref{tab1} reports the corresponding variable importance measures and the $p$-values, for 10 mutations that were also reported by \cite{bellot2019conditional}. We see that, the $p$-values of the tests generally agree well with the variable important measures from the EN and RF models. Meanwhile, the two conditional independence tests agree relatively well, except for two genetic mutations, MAP3K5 and FTL3. GCIT concluded that MAP3K5 is significant ($p < 0.001$) but FTL3 is not ($p = 0.521$), whereas our test leads to the opposite conclusion that MAP3K5 is insignificant ($p = 0.794$) but FTL3 is ($p = 0$). Besides, both EN and RF place FTL3 as an important mutation. We then compare our findings with the cancer drug response literature. Actually, MAP3K5 has not been previously reported in the literature as being directly linked to the PLX4720 drug response. Meanwhile, there is strong evidence showing the connections of the FLT3 mutation with cancer response \citep{Tsai2008, Larrosa-Garcia2017}. Combining the existing literature with our theoretical and synthetic results, we have more confidence about the findings of our proposed test.

\section{Discussion}
\label{sec:dis}

\change{In this article, we have developed a new inferential procedure for high-dimensional conditional independence testing, where the dimension of the conditional variables can diverge with the sample size. Our proposal utilizes a set of state-of-the-art deep learning tools to help address a classical statistics and machine learning problem. It integrates GANs, neural networks, cross-fitting and multiplier bootstrap. It achieves the asymptotic guarantees under much weaker conditions, and enjoys better empirical performances, when compared to the existing tests. As a tradeoff, our test is computationally more complicated. Nevertheless, the wall clock time for running the entire test for one data replication is in the order of a few minutes and is deemed reasonable. Finally, the computer code is publicly available on the GitHub repository: \url{https://github.com/tianlinxu312/dgcit}. }

\appendix
\section{Proofs}
We provide the proofs of Proposition \ref{prop1}, Theorems \ref{thm2}, \ref{thm3}, and \ref{thm4}. We omit the proof of Theorem \ref{thm1}, since it is similar to that of Theorem \ref{thm2}. We note that Theorems \ref{thm1}-\ref{thm4} are established under our choice of the function classes $\mathbb{H}_1$ and $\mathbb{H}_2$, which are set to the classes of neural networks with a single-hidden layer, finitely many hidden nodes, and the sigmoid activation function, as used in our implementation. Meanwhile, our results can be extended to more general choices of the function classes.

\subsection{Proof of Proposition \ref{prop1}}
\label{sec:proofprop1}

Note that the total variation distance is bounded by $1$. Suppose $\Mean d_{\scriptsize{\hbox{TV}}}(\widetilde{P}_{\bm{X}|\bm{Z}} ,P_{\bm{X}|\bm{Z}})=o(1)$. Then we have $d_{\scriptsize{\hbox{TV}}}(\widetilde{P}_{\bm{X}|\bm{Z}} ,P_{\bm{X}|\bm{Z}})=o_p(1)$. By the dominated convergence theorem, we have $\Mean d_{\scriptsize{\hbox{TV}}}^2(\widetilde{P}_{\bm{X}|\bm{Z}} ,P_{\bm{X}|\bm{Z}})=o(1)$. 

By Theorem 1.2 of \cite{devroye2018total}, we have $d_{\scriptsize{\hbox{TV}}}(\widetilde{P}_{\bm{X}|\bm{Z}} ,P_{\bm{X}|\bm{Z}})$ is proportional to
\begin{eqnarray*}
	\min\left[1,\sigma_0^{-1} \sqrt{\sum_{i=1}^n \left\{ Z_i^\top (\widehat{\beta}-\beta_0) \right\}^2} \right].
\end{eqnarray*}
It follows that
\begin{eqnarray*}
	\frac{1}{\sigma_0}\Mean \sum_{i=1}^n \{Z_i^\top (\widehat{\beta}-\beta_0)\}^2=o(1).
\end{eqnarray*}
Applying Theorem 1.2 of \cite{devroye2018total} again, we obtain that $d_{\scriptsize{\hbox{TV}}}(\widetilde{P}_{X|Z=Z_i} ,P_{X|Z=Z_i})$ is proportional to
\begin{eqnarray*}
	\min\left\{ 1,\sigma_0^{-1} | Z_i^\top (\widehat{\beta}-\beta_0) | \right\}.
\end{eqnarray*}
Therefore, we obtain that, 
\begin{eqnarray*}
	\sum_{i=1}^n \Mean d_{\scriptsize{\hbox{TV}}}^2\left( \widetilde{P}_{X|Z=Z_i} ,P_{X|Z=Z_i} \right) = o(1).
\end{eqnarray*}
Since the data is exchangeable, we have that, 
\begin{eqnarray}\label{eqn:proofprop1}
\Mean d_{\scriptsize{\hbox{TV}}}^2\left( \widetilde{P}_{X|Z=Z_i} ,P_{X|Z=Z_i} \right) = o(n^{-1}).
\end{eqnarray}
This shows that when RHS of \eqref{eqn:excesspvalue}, i.e., $\Mean \{ d_{\scriptsize{\hbox{TV}}}( \widetilde{P}_{\bm{X}|\bm{Z}} ,P_{\bm{X}|\bm{Z}} ) \}$ is $o(1)$, \eqref{eqn:proofprop1} holds. 

Next, we show (\ref{eqn:proofprop1}) is violated in the linear regression example. By the data exchangeability, it suffices to show $\sum_{i=1}^n \Mean d_{\scriptsize{\hbox{TV}}}^2\{\widetilde{P}_{X|Z=Z_i} ,P_{X|Z=Z_i}\}$ is not $o(1)$. With some calculations, we obtain that, 
\begin{align}\label{eqn:proofprop11}
\begin{split}	
& \sum_{i=1}^n \Mean \min\left\{1,\sigma_0^{-2} | Z_i^\top (\widehat{\beta}-\beta_0) |^2 \right\} \\ 
= \; & \sum_{i=1}^n \Mean\sigma_0^{-2} | Z_i^\top (\widehat{\beta}-\beta_0) |^2 \mathbb{I}\left\{ \sigma_0^{-2} |Z_i^\top (\widehat{\beta}-\beta_0)|^2\le 1\right\} + \sum_{i=1}^n \Mean \mathbb{I}\left\{ \sigma_0^{-2} |Z_i^\top (\widehat{\beta}-\beta_0)|^2> 1\right\} \\
= \; & \sum_{i=1}^n \Mean\sigma_0^{-2} |Z_i^\top (\widehat{\beta}-\beta_0)|^2 - \sum_{i=1}^n \Mean \left\{ \sigma_0^{-2} |Z_i^\top (\widehat{\beta}-\beta_0)|^2-1 \right\} \mathbb{I}\left\{ \sigma_0^{-2} |Z_i^\top (\widehat{\beta}-\beta_0)|^2> 1 \right\}.
\end{split}	
\end{align}
By the definition of $\widehat{\beta}$, we have
\begin{eqnarray*}
	\sum_{i=1}^n \Mean\sigma_0^{-2} |Z_i^\top (\widehat{\beta}-\beta_0)|^2=\frac{1}{\sigma_0^2} \Mean (\widehat{\beta}-\beta)^\top \bm{Z}^\top \bm{Z} (\widehat{\beta}-\beta)
	= \frac{1}{\sigma_0^2}\Mean  \bm{\varepsilon}^\top \bm{Z} (\bm{Z}^\top \bm{Z})^{-1} \bm{Z}^\top \bm{\varepsilon},
\end{eqnarray*}
where $\bm{\varepsilon}=(\varepsilon_1,\cdots,\varepsilon_n)^\top$ consist of i.i.d. copies of $\varepsilon$ defined in Example 1. It follows that, 
\begin{eqnarray}\label{eqn:proofprop115}
\begin{split}
\sum_{i=1}^n \Mean\sigma_0^{-2} |Z_i^\top (\widehat{\beta}-\beta_0)|^2=\frac{1}{\sigma_0^2}\Mean  \bm{\varepsilon}^\top \bm{Z} (\bm{Z}^\top \bm{Z})^\top \bm{Z}^\top \bm{\varepsilon}=\frac{1}{\sigma_0^2} \hbox{trace}\left\{\Mean \bm{\varepsilon} \bm{\varepsilon}^\top \bm{Z} (\bm{Z}^\top \bm{Z})^{-1}  \bm{Z}^\top \right\}\\
=\hbox{trace}\left\{\Mean\bm{Z} (\bm{Z}^\top \bm{Z})^{-1}  \bm{Z}^\top \right\}=d_Z,
\end{split}	
\end{eqnarray} 
where $d_Z$ is the dimension of $Z$.

Next, we show that, 
\begin{eqnarray}\label{eqn:proofprop12}
\sum_{i=1}^n \Mean \sigma_0^{-2} |Z_i^\top (\widehat{\beta}-\beta_0)|^2 \mathbb{I}\left\{ \sigma_0^{-2} |Z_i^\top (\widehat{\beta}-\beta_0)|^2\ge 1 \right\} = o(1),
\end{eqnarray}
or equivalently,
\begin{eqnarray*}
	\Mean n\sigma_0^{-2} |Z_i^\top (\widehat{\beta}-\beta_0)|^2 \mathbb{I}\left\{ \sigma_0^{-2} |Z_i^\top (\widehat{\beta}-\beta_0)|^2 \ge 1 \right\} = o(1).
\end{eqnarray*}
We have already shown that $\Mean n\sigma_0^{-2} |Z_i^\top (\widehat{\beta}-\beta_0)|^2=d_Z$. By the dominated convergence theorem, it suffices to show that, 
\begin{eqnarray*}
	n \sigma_0^{-2} |Z_i^\top (\widehat{\beta}-\beta_0)|^2 \mathbb{I}\left\{ \sigma_0^{-2} |Z_i^\top (\widehat{\beta}-\beta_0)|^2 \ge 1 \right\}=o_p(1).
\end{eqnarray*}
By definition, it in turn suffices to show that, 
\begin{eqnarray*}
	\prob\left\{ \sigma_0^{-2} |Z_i^\top (\widehat{\beta}-\beta_0)|^2\ge 1\right\} \to 0.
\end{eqnarray*}
This holds by Markov's inequality, as
\begin{eqnarray*}
	\Mean \sigma_0^{-2} |Z_i^\top (\widehat{\beta}-\beta_0)|^2=\frac{d_Z}{n}\to 0.
\end{eqnarray*}

Combining \eqref{eqn:proofprop12} together with \eqref{eqn:proofprop11} and \eqref{eqn:proofprop115} yields that, 
\begin{eqnarray*}
	\sum_{i=1}^n \Mean \min\left\{ 1,\sigma_0^{-2} |Z_i^\top (\widehat{\beta}-\beta_0)|^2 \right\} \ge d_Z-o(1) \ge 1-o(1),
\end{eqnarray*}
and hence $\sum_{i=1}^n \Mean d_{\scriptsize{\hbox{TV}}}^2\{\widetilde{P}_{X|Z=Z_i} ,Q_X^{(n)}(\cdot|Z_i)\}\ge 1-o(1)$. 

This completes the proof of Proposition \ref{prop1}. 
\eop
\bigskip

\subsection{Proof of Theorem \ref{thm2}}
\label{sec:proofthm2}

\change{We begin by providing an upper bound for the function classes $\mathbb{H}_1$ and $\mathbb{H}_2$. Recall that both $\mathbb{H}_1$ and $\mathbb{H}_2$ are classes of neural networks with a single-hidden layer, finitely many hidden nodes, and the sigmoid activation function. Because of that, each function $h_{1,\theta_1} \in \mathbb{H}_1$ and $h_{2,\theta_2} \in \mathbb{H}_2$ can be represented as
\begin{eqnarray*}
h_{1,\theta_1}(x)=\sum_{j=1}^M \theta_{1,j}^{(1)}\hbox{sigmoid}(x^\top \theta_{1,j}^{(2)}),\,\,\,\,h_{2,\theta_2}(x)=\sum_{j=1}^M \theta_{2,j}^{(1)}\hbox{sigmoid}(y^\top \theta_{2,j}^{(2)}),
\end{eqnarray*}
where $\theta_1$ and $\theta_2$ correspond to the sets of parameters $\big\{ (\theta_{1,j}^{(1)}, \theta_{1,j}^{(2)}) : 1\le j\le M \big\}$ and $\big\{ (\theta_{2,j}^{(1)}, \theta_{2,j}^{(2)}) : 1\le j\le M \big\}$, respectively, and $M$ is a finite integer. Note that the sigmoid function is bounded. As such, the functions $h_{1,\theta_1}$ and $h_{2,\theta_2}$ are uniformly bounded by $\sum_{j=1}^M |\theta_{1,j}^{(1)}|$ and $\sum_{j=1}^M |\theta_{2,j}^{(2)}|$, respectively. Since we sample $B$ many functions $\{h_{1,\theta_b}\}_{b=1}^B$ and $\{h_{2,\theta_b}\}_{b=1}^B$, these functions are uniformly bounded by
\begin{eqnarray*}
M \max_{b,j} \left( |\theta_{b,j}^{(1)}| + |\theta_{b,j}^{(2)}| \right). 
\end{eqnarray*}
Since these parameters $\theta_1, \theta_2$ are sampled from standard normal distributions, and that
\begin{eqnarray*}
\prob(W>t)=\frac{1}{\sqrt{2\pi}}\int_{t}^{\infty} \exp\left(-\frac{w^2}{2}\right)dw \le \frac{1}{\sqrt{2\pi}} \int_{t}^{\infty} w \exp\left(-\frac{w^2}{2}\right)dw = \frac{\exp(-t^2/2)}{\sqrt{2\pi}},
\end{eqnarray*}
for any $t \ge 1$, we can show that $\max_{b,j} \left( |\theta_{b,j}^{(1)}| + |\theta_{b,j}^{(2)}| \right)$ is upper bounded by $\sqrt{\log B}$, with probability approaching one. Note that $B$ grows polynomially with respect to the sample size $n$. Therefore, we have that the functions in $\mathbb{H}_1$ and $\mathbb{H}_2$ are upper bounded by $\log n$ in absolute values.} 

Define a test statistic
\begin{eqnarray*}
	T^{**}=\max_{b_1,b_2} \widehat{\sigma}_{b_1,b_2}^{-1} \left|\frac{1}{n}\sum_{i=1}^n \left\{ h_{1,b_1}(X_i) -\frac{1}{M}\sum_{m=1}^M h_{1,b_1}(X_i^{(m)}) \right\} \left\{ h_{2,b_2}(Y_i)-\frac{1}{M}\sum_{m=1}^M h_{2,b_2}(Y_i^{(m)}) \right\} \right|,
\end{eqnarray*}
where the $\widehat{\sigma}_{b_1,b_2}$ is constructed based on $\{\widetilde{X}_i^{(m)}\}_m$ and $\{\widetilde{Y}_i^{(m)}\}_m$, instead of $\{X_i^{(m)}\}_m$ and $\{Y_i^{(m)}\}_m$. It suffices to show that $|\widehat{T}-T^{**}|=O_p(n^{-2\kappa}\log n)$, and $|T^*-T^{**}|=O_p(n^{-2\kappa} \log n)$.

\medskip
\noindent
\textbf{Step 1.} 
We first consider the difference $|\widehat{T}-T^{**}|$. For any sequences $\{a_n\}_n$, $\{b_n\}_n$, we have that, 
\begin{eqnarray}\label{eqn:max}
|\max_n |a_n|-\max_n |b_n||\le \max_n |a_n-b_n|. 
\end{eqnarray}
Consequently, we have $|\widehat{T}-T^{**}| \le I_1+I_2+I_3$, where
\begin{eqnarray*}
	I_1 & = & \max_{b_1,b_2} \widehat{\sigma}_{b_1,b_2}^{-1} \left|\frac{1}{n}\sum_{i=1}^n \left[\frac{1}{M}\sum_{m=1}^M \left\{h_{1,b_1}(X_i^{(m)})-h_{1,b_1}(\widetilde{X}_i^{(m)}) \right\} \right] \left\{ h_{2,b_2}(Y_i) -\frac{1}{M}\sum_{m=1}^M h_{2,b_2}(Y_i^{(m)}) \right\}\right|,\\
	I_2 & = & \max_{b_1,b_2} \widehat{\sigma}_{b_1,b_2}^{-1} \left|\frac{1}{n}\sum_{i=1}^n \left\{ h_{1,b_1}(X_i) -\frac{1}{M}\sum_{m=1}^M h_{1,b_1}(X_i^{(m)}) \right\} \left[ \frac{1}{M}\sum_{m=1}^M \left\{ h_{2,b_2}(Y_i^{(m)})-h_{2,b_2}(\widetilde{Y}_i^{(m)}) \right\} \right]\right|,\\
	I_3 & = & \max_{b_1,b_2}\widehat{\sigma}_{b_1,b_2}^{-1} \left|\frac{1}{n}\sum_{i=1}^n \left[ \frac{1}{M}\sum_{m=1}^M \left\{ h_{1,b_1}(X_i^{(m)})-h_{1,b_1}(\widetilde{X}_i^{(m)}) \right\} \right]\left[ \frac{1}{M}\sum_{m=1}^M \left\{ h_{2,b_2}(Y_i^{(m)})-h_{2,b_2}(\widetilde{Y}_i^{(m)}) \right\} \right]\right|.
\end{eqnarray*}
If $\min \widehat{\sigma}_{b_1,b_2}\ge c_0$ for some constant $c_0>0$, then it suffices to show that $I_j^*=O_p(n^{-(\kappa_x+\kappa_y)}\log n)$, for $j=1,2,3$, where 
\begin{eqnarray*}
	I_1^* & = & \max_{b_1,b_2} \left|\frac{1}{n}\sum_{i=1}^n \left[\frac{1}{M}\sum_{m=1}^M \left\{ h_{1,b_1}(X_i^{(m)})-h_{1,b_1}(\widetilde{X}_i^{(m)}) \right\} \right] \left\{ h_{2,b_2}(Y_i) -\frac{1}{M}\sum_{m=1}^M h_{2,b_2}(Y_i^{(m)}) \right\} \right|,\\
	I_2^* & = & \max_{b_1,b_2} \left|\frac{1}{n}\sum_{i=1}^n \left\{ h_{1,b_1}(X_i) -\frac{1}{M}\sum_{m=1}^M h_{1,b_1}(X_i^{(m)}) \right\}\left[ \frac{1}{M}\sum_{m=1}^M \left\{ h_{2,b_2}(Y_i^{(m)})-h_{2,b_2}(\widetilde{Y}_i^{(m)}) \right\} \right]\right|,\\
	I_3^* & = & \max_{b_1,b_2} \left|\frac{1}{n}\sum_{i=1}^n \left[\frac{1}{M}\sum_{m=1}^M \left\{ h_{1,b_1}(X_i^{(m)})-h_{1,b_1}(\widetilde{X}_i^{(m)}) \right\} \right] \left[ \frac{1}{M}\sum_{m=1}^M \left\{ h_{2,b_2}(Y_i^{(m)})-h_{2,b_2}(\widetilde{Y}_i^{(m)}) \right\} \right]\right|.
\end{eqnarray*}
The number of folds $L$ is finite, as such, it suffices to show that $I_j^{(\ell)}=O_p(n^{-(\kappa_x+\kappa_y)}\log n)$, for $j=1,2,3$ and $\ell=1,\ldots,L$, where
\begin{eqnarray*}
	I_1^{(\ell)} & = & \max_{b_1,b_2} \left|\frac{1}{n}\sum_{i\in \mathcal{I}^{(\ell)}} \left[\frac{1}{M}\sum_{m=1}^M \left\{ h_{1,b_1}(X_i^{(m)})-h_{1,b_1}(\widetilde{X}_i^{(m)}) \right\} \right] \left\{ h_{2,b_2}(Y_i) -\frac{1}{M}\sum_{m=1}^M h_{2,b_2}(Y_i^{(m)}) \right\}\right|,\\
	I_2^{(\ell)} & = & \max_{b_1,b_2} \left|\frac{1}{n}\sum_{i\in \mathcal{I}^{(\ell)}} \left\{ h_{1,b_1}(X_i) -\frac{1}{M}\sum_{m=1}^M h_{1,b_1}(X_i^{(m)}) \right\} \left[ \frac{1}{M}\sum_{m=1}^M \left\{ h_{2,b_2}(Y_i^{(m)})-h_{2,b_2}(\widetilde{Y}_i^{(m)}) \right\} \right]\right|,\\
	I_3^{(\ell)} & = & \max_{b_1,b_2} \left|\frac{1}{n}\sum_{i\in \mathcal{I}^{(\ell)}} \left[\frac{1}{M}\sum_{m=1}^M \left\{ h_{1,b_1}(X_i^{(m)})-h_{1,b_1}(\widetilde{X}_i^{(m)}) \right\} \right] \left[ \frac{1}{M}\sum_{m=1}^M \left\{ h_{2,b_2}(Y_i^{(m)})-h_{2,b_2}(\widetilde{Y}_i^{(m)}) \right\} \right]\right|.
\end{eqnarray*}
We divide the rest of the proof into four sub-steps. We first show that $I_j^{(\ell)}=O_p(n^{-(\kappa_x+\kappa_y)}\log n)$, for $j=1,2,3$. Finally, we show $\prob(\min\widehat{\sigma}_{b_1,b_2}\ge c_0)\to 1$ for some constant $c_0>0$.

\medskip
\noindent
\textbf{Step 1.1.} 
\change{Recall we have shown that the functions in $\mathbb{H}_1$ and $\mathbb{H}_2$ are bounded by $\log n$ in absolute values at the beginning of the proof of Theorem \ref{thm2}.} By Bernstein's inequality, we have that, 
\begin{eqnarray*}
	\prob\left[\left|\sum_{m=1}^M h_{1,b}(X_i^{(m)})- M\Mean\{ h_{1,b}(X_i)|Z_i\}\right|\ge t\right] \le 2\exp\left\{-\frac{t^2}{2(M\log n+t\sqrt{\log n}/3)} \right\},
\end{eqnarray*}
for any $b$ and $i$. Set $t= \sqrt{3(c+2)M} \log n$, where the constant $c$ is as defined in the statement of Theorem \ref{thm1}. For a sufficiently large $n$, we have $t\sqrt{\log n}/3\le M\log n/2$. It follows that
\begin{eqnarray*}
	\prob\left[\left|\sum_{m=1}^M h_{1,b}(X_i^{(m)})- M\Mean\{ h_{1,b}(X_i)|Z_i\}\right| \ge \sqrt{3(c+2)M}\log n\right] \le \frac{2}{n^{c+2}}.
\end{eqnarray*}
By Bonferroni's inequality, we obtain that, 
\begin{eqnarray*}
	&&\prob\left[ \max_{b\in \{1,\cdots,B\}} \max_{i\in \{1,\cdots,n\}} \left|\sum_{m=1}^M h_{1,b}(X_i^{(m)})- M\Mean\{ h_{1,b}(X_i)|Z_i\}\right|\ge \sqrt{3(c+2)M}\log n\right] \\ 
	&\le& Bn \max_{b\in \{1,\cdots,B\}} \max_{i\in \{1,\cdots,n\}} \prob\left[ \left|\sum_{m=1}^M h_{1,b}(X_i^{(m)})-M \Mean\{ h_{1,b}(X_i)|Z_i\}\right|\ge \sqrt{3(c+2)M}\log n \right] \le \frac{2Bn}{n^{c+2}}.	
\end{eqnarray*}
Under the condition $B=O(n^c)$, we obtain with probability $1-O(n^{-1})$ that,
\begin{eqnarray}\label{eqn:event1}
\max_{b\in \{1,\cdots,B\}} \max_{i\in \{1,\cdots,n\}} \left|\sum_{m=1}^M h_{1,b}(X_i^{(m)})- M\Mean\{ h_{1,b}(X_i)|Z_i\}\right|\le O(1)n^{-1/2}\log n,
\end{eqnarray}
as $M$ is proportional to $n$, and $O(1)$ denotes some positive constant. 

Similarly, we can show that, 
\begin{eqnarray*}
	\max_{b\in \{1,\cdots,B\}} \max_{i\in \mathcal{I}^{(\ell)}} \left|\sum_{m=1}^M h_{1,b}(\widetilde{X}_i^{(m)})- M\int_x h_{1,b}(x)\widetilde{P}_{X|Z=Z_i}^{(\ell)}(dx) \right|\le O(1)\sqrt{n}\log n,
\end{eqnarray*}
with probability $1-O(n^{-1})$. Combining this with \eqref{eqn:event1}, we obtain with probability $1-O(n^{-1})$ that, 
\begin{align}\label{eqn:event2}
\begin{split}
\max_{\substack{b\in \{1,\ldots,B\}\\ i\in \mathcal{I}^{(\ell)} }} \left| \sum_{m=1}^M \right. & \left\{ h_{1,b}(X_i^{(m)})-h_{1,b}(\widetilde{X}_i^{(m)}) \right\} \\
& \left. - M\int_x h_{1,b}(x) \left\{ P_{X|Z=Z_i}(dx)-\widetilde{P}_{X|Z=Z_i}^{(\ell)}(dx) \right\} \right|
\le O(1)\sqrt{n}\log n.
\end{split}
\end{align}

Conditional on $Z_i$, the expectation of $h_{2,b_2}(Y_i)-M^{-1} \sum_{m=1}^M h_{2,b_2}(Y_i^{(m)})$ equals zero. Under the null hypothesis, the expectation of $M^{-1}\sum_{m=1}^M \{h_{1,b_1}(X_i^{(m)})-h_{1,b_1}(\widetilde{X}_i^{(m)})\}\{h_{2,b_2}(Y_i)-M^{-1} \sum_{m=1}^M h_{2,b_2}(Y_i^{(m)})\}$ equals zero as well. Applying Bernstein's inequality again, we can show with probability tending to $1$ that,
\begin{eqnarray}\label{eqn:inequality1}
I_1^{(\ell)} \le O(1) \left( \sigma n^{-1/2}\log^{3/2} n+n^{-1}\log^2 n \right),
\end{eqnarray}
where 
\begin{eqnarray*}
	\sigma^2=\max_{b_1,b_2} \Mean \left|\frac{1}{M}\sum_{m=1}^M \left\{ h_{1,b_1}(X_i^{(m)})-h_{1,b_1}(\widetilde{X}_i^{(m)}) \right\} \left\{h_{2,b_2}(Y_i)- \frac{1}{M}\sum_{m=1}^M h_{2,b_2}(Y_i^{(m)}) \right\}\right|^2\\
	\le \max_{b_1} \Mean \left|\frac{1}{M}\sum_{m=1}^M \left\{ h_{1,b_1}(X_i^{(m)})-h_{1,b_1}(\widetilde{X}_i^{(m)}) \right\}\right|^2\log n.
\end{eqnarray*}
Let $\mathcal{A}$ denote the event in \eqref{eqn:event2}. The last term on the second line can be bounded from above by
\begin{eqnarray}\label{eqn:firstline}
&&\max_{b_1,i} \Mean \left|\frac{1}{M}\sum_{m=1}^M \{h_{1,b_1}(X_i^{(m)})-h_{1,b_1}(\widetilde{X}_i^{(m)})\}\right|^2\mathbb{I}(\mathcal{A})\log n\\\label{eqn:secondline}
&+&\max_{b_1,i} \Mean \left|\frac{1}{M}\sum_{m=1}^M \{h_{1,b_1}(X_i^{(m)})-h_{1,b_1}(\widetilde{X}_i^{(m)})\}\right|^2\mathbb{I}(\mathcal{A}^c)\log n.
\end{eqnarray}
Since $M$ is proportional to $n$, by \eqref{eqn:max}, \eqref{eqn:firstline} is upper bounded by
\begin{eqnarray*}
	O(1) \left[ n^{-1}\log^2 n+\max_{\substack{b\in \{1,\cdots,B\}\\ i\in \mathcal{I}^{(\ell)}}}\Mean \left|\int_x h_{1,b}(x) \left\{ \widetilde{P}_{X|Z=Z_i}^{(\ell)}(dx)-P_{X|Z=Z_i}(dx) \right\}\right|^2 \right] \log n.
\end{eqnarray*}
By the boundedness of the function class $\mathbb{H}_1$, it can be further bounded from above by
\begin{eqnarray}\label{eqn:inequality2}
O(1) \left\{n^{-1}\log^3 n+\Mean d_{\scriptsize{\hbox{TV}}}^2(\widetilde{P}^{(\ell)}_{X|Z} ,P_{X|Z})\log^2 n \right\}.
\end{eqnarray}
The above quantity is of order $O(n^{-2\kappa_x}\log^2 n)$. Consequently, \eqref{eqn:firstline} is of the order $O(n^{-2\kappa_x}\log^2 n)$. 

Note that the event $\mathcal{A}$ occurs with probability at least $1-O(n^{-1})$. By the boundedness of the function class $\mathbb{H}_1$, \eqref{eqn:secondline} is of the order $O(n^{-1}\log^2 n)$. 

Therefore, $\sigma^2$ is of the order $O(n^{-2\kappa_x}\log^2 n)$. This implies that $\mathcal{I}_1^{(\ell)}$ can be bounded from above by $O(n^{-1/2-\kappa_x}\log^{5/2} n)$, which in turn yields that $\mathcal{I}_1^{(\ell)}=O_p(n^{-\kappa_x-\kappa_y}\log n)$, since $\kappa_x,\kappa_y<1/2$.

\medskip
\noindent
\textbf{Step 1.2.} This step can be proven in a similar way as Step 1.1, and is omitted.

\medskip
\noindent
\textbf{Step 1.3.} 
Under $H_0$, the expectation of 
\begin{eqnarray*}
	\frac{1}{|\mathcal{I}^{(\ell)}|}\sum_{i\in \mathcal{I}^{(\ell)}} \left[\frac{1}{M}\sum_{m=1}^M \left\{ h_{1,b_1}(X_i^{(m)})-h_{1,b_1}(\widetilde{X}_i^{(m)}) \right\} \right]\left[ \frac{1}{M}\sum_{m=1}^M \left\{ h_{2,b_2}(Y_i^{(m)})-h_{2,b_2}(\widetilde{Y}_i^{(m)}) \right\} \right]
\end{eqnarray*}
equals 
\begin{eqnarray*}
	\Mean \int_x h_{1,b_1}(x) \left\{ \widetilde{P}^{(\ell)}_{X|Z}(dx)-P_{X|Z}(dx) \right\} \int_y h_{2,b_2}(y)\left\{ \widetilde{P}^{(\ell)}_{Y|Z}(dy)-P_{Y|Z}(dy) \right\}.
\end{eqnarray*}
Similar to \eqref{eqn:inequality2}, its absolute value can be upper bounded by
\begin{eqnarray*}
	\Mean d_{\scriptsize{\hbox{TV}}}\left\{ \widetilde{P}^{(\ell)}_{X|Z=Z_i} ,P_{X|Z} \right\} d_{\scriptsize{\hbox{TV}}}\left\{ \widetilde{P}^{(\ell)}_{Y|Z=Z_i} ,P_{Y|Z} \right\} \log n.
\end{eqnarray*}
Following Cauchy-Schwarz inequality, we have that, 
\begin{eqnarray*}
&&\Mean d_{\scriptsize{\hbox{TV}}}\left\{ \widetilde{P}^{(\ell)}_{X|Z=Z_i} ,P_{X|Z} \right\} d_{\scriptsize{\hbox{TV}}}\left\{ \widetilde{P}^{(\ell)}_{Y|Z=Z_i} ,P_{Y|Z} \right\} \\
&\le& \sqrt{\Mean d_{\scriptsize{\hbox{TV}}}^2\left\{ \widetilde{P}^{(\ell)}_{X|Z=Z_i} ,P_{X|Z} \right\}\Mean d_{\scriptsize{\hbox{TV}}}^2\left\{ \widetilde{P}^{(\ell)}_{Y|Z=Z_i} ,P_{Y|Z} \right\}}=O(n^{-(\kappa_x+\kappa_y)}).
\end{eqnarray*}
This yields that,
\begin{eqnarray*}
	\max_{b_1,b_2}\left|\Mean \int_x h_{1,b_1}(x)\left\{ \widetilde{P}^{(\ell)}_{X|Z}(dx)-P_{X|Z}(dx) \right\} \int_y h_{2,b_2}(y)\left\{ \widetilde{P}^{(\ell)}_{Y|Z}(dy)-P_{Y|Z}(dy) \right\}\right| = O(n^{-(\kappa_x+\kappa_y)}\log n). 
\end{eqnarray*}
Following similar arguments as in Step 1.1, we obtain that,
\begin{eqnarray*}
	I_3^{(\ell)} - \max_{b_1,b_2}\left|\Mean \int_x h_{1,b_1}(x)\left\{ \widetilde{P}^{(\ell)}_{X|Z}(dx)-P_{X|Z}(dx) \right\} \int_y h_{2,b_2}(y)\left\{ \widetilde{P}^{(\ell)}_{Y|Z}(dy)-P_{Y|Z}(dy) \right\}\right|\\ = O_p(n^{-(\kappa_x+\kappa_y)}\log n).
\end{eqnarray*}
Therefore, we obtain that $I_3^{(\ell)}=O_p(n^{-(\kappa_x+\kappa_y)}\log n)$.

\medskip
\noindent
\textbf{Step 1.4.} 
Recall that $\widehat{\sigma}_{b_1,b_2}^2$ is defined by
\begin{eqnarray*}
	\frac{1}{n-1} \sum_{i=1}^n \bigg( \left[ h_{1,b_1}(X_i) -\widehat{\Mean }\{h_{1,b_1}(X_i)|Z_i\} \right] \left[ h_{2,b_2}(Y_i) -\widehat{\Mean }\{h_{2,b_2}(Y_i)|Z_i\} \right]- \hbox{GCM}\{ h_{1,b_1}(X),h_{2,b_2}(Y) \} \bigg)^2.
\end{eqnarray*}
With some calculations, it is equal to
\begin{eqnarray}\label{eqn:sigma2}
\begin{split}
\frac{1}{n-1} \sum_{i=1}^n \left[ h_{1,b_1}(X_i) -\widehat{\Mean }\{h_{1,b_1}(X_i)|Z_i\} \right]^2 \left[ h_{2,b_2}(Y_i) -\widehat{\Mean }\{h_{2,b_2}(Y_i)|Z_i\} \right]^2\\
-\frac{n}{n-1}\hbox{GCM}^2\{ h_{1,b_1}(X),h_{2,b_2}(Y) \},
\end{split}	
\end{eqnarray}
where the estimated conditional expectation $\widehat{\mathbb{E}}$ is computed using GANs. 

Consider the second term $\hbox{GCM}\{ h_{1,b_1}(X),h_{2,b_2}(Y) \}$ in \eqref{eqn:sigma2}. Following similar arguments as in Steps 1.1 and 1.3, we have that, 
\begin{eqnarray*}
	\max_{b_1,b_2} \left| \hbox{GCM}\{ h_{1,b_1}(X),h_{2,b_2}(Y) \}-\hbox{GCM}'\{ h_{1,b_1}(X),h_{2,b_2}(Y) \} \right| = O_p(n^{-(\kappa_x+\kappa_y)}\log n),
\end{eqnarray*}
where $\hbox{GCM}'\{ h_{1,b_1}(X),h_{2,b_2}(Y) \}$ equals 
\begin{eqnarray*}
	\frac{1}{n}\sum_{i=1}^n \left\{ h_{1,b_1}(X_i)-\frac{1}{M}\sum_{m=1}^M h_{1,b_1}(X_i^{(m)})\right\} \left\{ h_{2,b_2}(Y_i)-\frac{1}{M}\sum_{m=1}^M h_{2,b_2}(Y_i^{(m)}) \right\}.
\end{eqnarray*}
Similar to \eqref{eqn:event2}, we can show that, 
\begin{eqnarray*}
	\max_{b_1,b_2}\left| \hbox{GCM}'\{ h_{1,b_1}(X),h_{2,b_2}(Y) \}-\hbox{GCM}^*\{ h_{1,b_1}(X),h_{2,b_2}(Y) \} \right| = O_p\left( n^{-1/2}\sqrt{\log n} \right).
\end{eqnarray*} 
Consequently, we have that, 
\begin{eqnarray*}
	\max_{b_1,b_2} \left| \hbox{GCM}\{ h_{1,b_1}(X),h_{2,b_2}(Y) \}-\hbox{GCM}^*\{ h_{1,b_1}(X),h_{2,b_2}(Y) \} \right| = O_p\left( n^{-1/2}\sqrt{\log n} \right).
\end{eqnarray*}
Since the function classes $\mathbb{H}_1$ and $\mathbb{H}_2$ are bounded, both GCM and GCM$^*$ are bounded by $\log n$ in absolute values. Consequently,
\begin{eqnarray}\label{eqn:sigma2eq1}
\max_{b_1,b_2}\left| \hbox{GCM}^2\{ h_{1,b_1}(X),h_{2,b_2}(Y) \}-\hbox{GCM}^{*2}\{h_{1,b_1}(X),h_{2,b_2}(Y)\} \right|= O_p\big( n^{-1/2}\log^{3/2} n \big).
\end{eqnarray}

Next, consider the first term in \eqref{eqn:sigma2}. Note that it can be represented by
\begin{eqnarray*}
	\frac{n}{n-1}\frac{1}{L}\sum_{\ell=1}^L \left(\frac{1}{|\mathcal{I}^{\ell}|}\sum_{i\in \mathcal{I}^{(\ell)}}\left[ h_{1,b_1}(X_i) -\widehat{\Mean }\{h_{1,b_1}(X_i)|Z_i\} \right]^2 \left[ h_{2,b_2}(Y_i) -\widehat{\Mean }\{h_{2,b_2}(Y_i)|Z_i\} \right]^2\right).
\end{eqnarray*}
Similar to \eqref{eqn:event2}, we can show that, 
\begin{align*}
\max_{b_1,b_2} & \left|\frac{1}{|\mathcal{I}^{\ell}|} \sum_{i\in \mathcal{I}^{(\ell)}} \left[ h_{1,b_1}(X_i) -\widehat{\Mean }\{h_{1,b_1}(X_i)|Z_i\} \right]^2 \left[ h_{2,b_2}(Y_i) -\widehat{\Mean }\{h_{2,b_2}(Y_i)|Z_i\} \right]^2 \right. \\
& - \left.\Mean \left[ h_{1,b_1}(X_1) -\widehat{\Mean }\{h_{1,b_1}(X_1)|Z_1\} \right]^2 \left[ h_{2,b_2}(Y_1) -\widehat{\Mean }\{h_{2,b_2}(Y_1)|Z_1\} \right]^2\right|=O_p(n^{-1/2}\log^{3/2} n).
\end{align*}
Following similar arguments as in Steps 1.1 and 1.3, we can show that, 
\begin{align*}
\max_{b_1,b_2} & \left|\Mean \left[ h_{1,b_1}(X_1) -\widehat{\Mean }\{h_{1,b_1}(X_1)|Z_1\} \right]^2 \left[ h_{2,b_2}(Y_1) -\widehat{\Mean }\{h_{2,b_2}(Y_1)|Z_1\} \right]^2 \right.\\
& - \left.\Mean \left[ h_{1,b_1}(X_1) -\Mean\{h_{1,b_1}(X_1)|Z_1\} \right]^2 \left[ h_{2,b_2}(Y_1) -\Mean\{h_{2,b_2}(Y_1)|Z_1\} \right]^2\right|=O_p(n^{-\bar{c}}),
\end{align*}
for some constant $0<\bar{c}<1/2$. It follows that,
\begin{eqnarray*}
	\max_{b_1,b_2}\left|\frac{1}{|\mathcal{I}^{\ell}|}\sum_{i\in \mathcal{I}^{(\ell)}}\left[ h_{1,b_1}(X_i) -\widehat{\Mean }\{h_{1,b_1}(X_i)|Z_i\} \right]^2 \left[ h_{2,b_2}(Y_i) -\widehat{\Mean }\{h_{2,b_2}(Y_i)|Z_i\} \right]^2\right.\\
	-\left.\Mean \left[ h_{1,b_1}(X) -\Mean\{h_{1,b_1}(X)|Z\} \right]^2 \left[ h_{2,b_2}(Y) -\Mean\{h_{2,b_2}(Y)|Z\} \right]^2\right|=O_p(n^{-\bar{c}}),
\end{eqnarray*}
and henceforth, 
\begin{eqnarray*}
	\max_{b_1,b_2}\left|\frac{1}{n}\sum_{i=1}^n\left[ h_{1,b_1}(X_i) -\widehat{\Mean }\{h_{1,b_1}(X_i)|Z_i\} \right]^2 \left[ h_{2,b_2}(Y_i) -\widehat{\Mean }\{h_{2,b_2}(Y_i)|Z_i\} \right]^2\right.\\
	-\left.\Mean \left[ h_{1,b_1}(X) -\Mean\{h_{1,b_1}(X)|Z\} \right]^2 \left[ h_{2,b_2}(Y) -\Mean\{h_{2,b_2}(Y)|Z\} \right]^2\right|=O_p(n^{-\bar{c}}).
\end{eqnarray*}
Combining this together with \eqref{eqn:sigma2eq1} yields that, 
\begin{eqnarray*}
	\max_{b_1,b_2}\left|\widehat{\sigma}_{b_1,b_2}^2-\frac{n}{n-1} \Var\Big(\left[ h_{1,b_1}(X) -\Mean\{h_{1,b_1}(X)|Z\} \right] \left[ h_{2,b_2}(Y) -\Mean\{h_{2,b_2}(Y)|Z\} \right]\Big)\right|=O_p(n^{-\bar{c}}).
\end{eqnarray*}
Then, we have that, 
\begin{eqnarray*}
	\min_{b_1,b_2} \Var\Big(\left[ h_{1,b_1}(X) -\Mean\{h_{1,b_1}(X)|Z\} \right] \left[ h_{2,b_2}(Y) -\Mean\{h_{2,b_2}(Y)|Z\} \right]\Big)\ge c^*,
\end{eqnarray*}
for some constant $c^*>0$. Therefore, we have that
\begin{eqnarray*}
	\min_{b_1,b_2} \widehat{\sigma}_{b_1,b_2}^2\ge 2^{-1} c^*,
\end{eqnarray*}
with probability tending to $1$.

\medskip
\noindent
\textbf{Step 2.} We next consider the difference $|T^*-T^{**}|$, and show that it is of the order $O_p(n^{-2\kappa}\log n)$. Denote by $\widehat{\sigma}_{b_1,b_2}^{*2}$ the variance estimator with $\{\widetilde{X}_i^{(m)}\}_m$ and $\{\widetilde{Y}_i^{(m)}\}_m$ replaced by $\{X_i^{(m)}\}_m$ and $\{Y_i^{(m)}\}_m$. Using (\ref{eqn:max}), the difference between $T^*$ and $T^{**}$ is upper bounded by
\begin{eqnarray*}
	\max_{b_1,b_2} |\widehat{\sigma}_{b_1,b_2}^{-1}-\widehat{\sigma}_{b_1,b_2}^{*-1}|\left| \frac{1}{n} \sum_{i=1}^n \left\{ h_{1,b_1}(X_i)-\frac{1}{M}\sum_{m=1}^M h_{1,b_1}(X_i^{(m)}) \right\} \left\{ h_{2,b_2}(Y_i)-\frac{1}{M}\sum_{m=1}^M h_{2,b_2}(Y_i^{(m)}) \right\} \right|.
\end{eqnarray*}
Under $H_0$, similar to (\ref{eqn:event2}), we can show that,
\begin{eqnarray*}
	\max_{b_1,b_2} |\left| \frac{1}{n} \sum_{i=1}^n \left\{ h_{1,b_1}(X_i)-\frac{1}{M}\sum_{m=1}^M h_{1,b_1}(X_i^{(m)}) \right\} \left\{ h_{2,b_2}(Y_i)-\frac{1}{M}\sum_{m=1}^M h_{2,b_2}(Y_i^{(m)}) \right\} \right| \\
	=O_p(n^{-1/2}\log^{3/2} n). 
\end{eqnarray*}
To show $|T^*-T^{**}|=O_p(n^{-2\kappa}\log n)$, it suffices to show that $\max_{b_1,b_2} |\widehat{\sigma}_{b_1,b_2}^{-1}-\widehat{\sigma}_{b_1,b_2}^{*-1}|=O_p(n^{-\bar{c}})$ for some constant $\bar{c}>0$. Since both $\widehat{\sigma}_{b_1,b_2}^{-1}$ and $\widehat{\sigma}_{b_1,b_2}$ are bounded away from zero, it suffices to show that $\max_{b_1,b_2} |\widehat{\sigma}_{b_1,b_2}^2-\widehat{\sigma}_{b_1,b_2}^{*2}|=O_p(n^{-\bar{c}})$. 

Following similar arguments as in Steps 1.1 and 1.3, we can show that,
\begin{eqnarray*}
	&& \max_{b_1,b_2}\left|\widehat{\sigma}_{b_1,b_2}^2-\frac{n}{n-1} \Var\Big(\left[ h_{1,b_1}(X) -\Mean\{h_{1,b_1}(X)|Z\} \right] \left[ h_{2,b_2}(Y) -\Mean\{h_{2,b_2}(Y)|Z\} \right]\Big)\right|=O_p(n^{-\bar{c}}), \\
	&& \max_{b_1,b_2}\left|\widehat{\sigma}_{b_1,b_2}^{*2}-\frac{n}{n-1} \Var\Big(\left[ h_{1,b_1}(X) -\Mean\{h_{1,b_1}(X)|Z\} \right] \left[ h_{2,b_2}(Y) -\Mean\{h_{2,b_2}(Y)|Z\} \right]\Big)\right|=O_p(n^{-\bar{c}}).
\end{eqnarray*}
This completes the proof of Theorem \ref{thm2}.  
\eop
\bigskip

\subsection{Proof of Theorem \ref{thm3}}
\label{sec:proofthm3}

In the proof of Theorem \ref{thm2}, we have already shown that $\widehat{T}-T^{*}=O_p(n^{-(\kappa_x+\kappa_y)}\log n)$. Following similar arguments as in Step 1.4, we can show that $T^*-T^{***}=O_p(n^{-(\kappa_x+\kappa_y)}\log n)$, where 
\begin{eqnarray*}
	T^{***}=\max_{b_1,b_2} \sigma_{b_1,b_2}^{-1} |\left|n^{-1}\sum_{i=1}^n \left\{ h_{1,b_1}(X_i)-\frac{1}{M}\sum_{m=1}^M h_{1,b_1}(X_i^{(m)}) \right\} \left\{h_{2,b_2}(Y_i)-\frac{1}{M}\sum_{m=1}^M h_{2,b_2}(Y_i^{(m)})\right\} \right|,
\end{eqnarray*} 
where 
\begin{eqnarray*}
	\sigma_{b_1,b_2}^2=\frac{n}{n-1} \Var\Big(\left[ h_{1,b_1}(X) -\Mean\{h_{1,b_1}(X)|Z\} \right] \left[ h_{2,b_2}(Y) -\Mean\{h_{2,b_2}(Y)|Z\} \right]\Big).
\end{eqnarray*}
By \eqref{eqn:event1}, following similar arguments as in the proof regarding the term $I_1$ in Theorem \ref{thm2}, we can show that $T^{***}-T^{****}=O_p(n^{-(\kappa_x+\kappa_y)}\log n)$, where 
\begin{eqnarray*}
	T^{****}=\max_{b_1,b_2} \sigma_{b_1,b_2}^{-1} |\left|n^{-1}\sum_{i=1}^n \left[h_{1,b_1}(X_i)-\Mean \{h_{1,b_1}(X_i)|Z_i\} \right]\left\{h_{2,b_2}(Y_i)-\frac{1}{M}\sum_{m=1}^M h_{2,b_2}(Y_i^{(m)})\right\} \right|.
\end{eqnarray*} 
Similarly, we can show that $T^{****}-T_0=O_p(n^{-(\kappa_x+\kappa_y)}\log n)$, where
\begin{eqnarray*}
	T_0=\max_{b_1,b_2} \sigma_{b_1,b_2}^{-1} |\left|n^{-1}\sum_{i=1}^n \left[h_{1,b_1}(X_i)-\Mean \{h_{1,b_1}(X_i)|Z_i\} \right]\left[h_{2,b_2}(Y_i)-\Mean \{h_{2,b_2}(Y_i)|Z_i\}\right] \right|.
\end{eqnarray*}
Therefore, we have shown that $\widehat{T}-T_0=O_p(n^{-(\kappa_x+\kappa_y)}\log n)$. Since $\kappa_x+\kappa_y>1/2$, we have that, 
\begin{eqnarray}\label{eqn:thm2step0}
\sqrt{n}(\widehat{T}-T_0)=o_p(\log^{-1/2} n).
\end{eqnarray}
Define a $B^2\times B^2$ matrix $\Sigma_0$ whose $\{b_1+B(b_2-1),b_3+B(b_4-1)\}$th entry is given by 
\begin{eqnarray*}
	\Cov\left(\sigma_{b_1,b_2}^{-1} \left[h_{1,b_1}(X_i)-\Mean \{h_{1,b_1}(X_i)|Z_i\} \right]\left[h_{2,b_2}(Y_i)-\Mean \{h_{2,b_2}(Y_i)|Z_i\}\right],\right.\\
	\left. \sigma_{b_3,b_4}^{-1} \left[h_{1,b_3}(X_i)-\Mean \{h_{1,b_3}(X_i)|Z_i\} \right]\left[h_{2,b_4}(Y_i)-\Mean \{h_{2,b_4}(Y_i)|Z_i\}\right]\right).
\end{eqnarray*}
In the following, we show that, 
\begin{eqnarray}\label{eqn:thm2step1}
\sup_t \left| \prob\left( \sqrt{n}\widehat{T}_0\le t|\mathcal{H}_0 \right) - \prob\left( \|N(0,\Sigma_0)\|_{\infty}\le t \right) \right| = o(1).
\end{eqnarray}
When $B$ is finite, this is implied by the classical weak convergence results. When $B$ diverges with $n$, we require $B=O(n^c)$ for some constant $c>0$. By the definition of $\sigma_{b_1,b_2}$, the variance of 
\begin{eqnarray*}
	\sigma_{b_1,b_2}^{-1} \left[h_{1,b_1}(X_i)-\Mean \{h_{1,b_1}(X_i)|Z_i\} \right]\left[h_{2,b_2}(Y_i)-\Mean \{h_{2,b_2}(Y_i)|Z_i\}\right]
\end{eqnarray*}
is bounded from above by $(n-1)/n$. Moreover, combining the boundedness of the function spaces $\mathbb{H}_1$ and $\mathbb{H}_2$ together with the definition of $\sigma_{b_1,b_2}$ yields that,
\begin{eqnarray*}
	\left\{\sigma_{b_1,b_2}^{-1} \left[h_{1,b_1}(X_i)-\Mean \{h_{1,b_1}(X_i)|Z_i\} \right]\left[h_{2,b_2}(Y_i)-\Mean \{h_{2,b_2}(Y_i)|Z_i\}\right]:b_1,b_2\in \{1,\cdots,B\}\right\}
\end{eqnarray*}
are uniformly bounded from infinity by $O(\log n)$, with probability tending to $1$.  
We can show that \eqref{eqn:thm2step1} holds. This implies that, 
\begin{eqnarray*}
	\sigma_{b_1,b_2}^{-1} n^{-1/2}\sum_{i=1}^n \left[h_{1,b_1}(X_i)-\Mean \{h_{1,b_1}(X_i)|Z_i\} \right]\left[h_{2,b_2}(Y_i)-\Mean \{h_{2,b_2}(Y_i)|Z_i\}\right]
\end{eqnarray*}
is asymptotically normal with zero mean. 

Combining \eqref{eqn:thm2step1} together with \eqref{eqn:thm2step0} yields that, 
\begin{eqnarray}\label{eqn:thm2step2}
\begin{split}
\prob\left( \sqrt{n}\widehat{T}\le t|\mathcal{H}_0 \right) \ge \prob\left( \|N(0,\Sigma_0)\|_{\infty}\le t-\epsilon_0 \log^{-1/2} n \right)-o(1),\\
\prob\left( \sqrt{n}\widehat{T}\le t|\mathcal{H}_0 \right) \le \prob\left( \|N(0,\Sigma_0)\|_{\infty}\le t+\epsilon_0 \log^{-1/2} n \right)+o(1),
\end{split}	
\end{eqnarray}
for any sufficiently small $\epsilon_0>0$, where the little-o terms are uniform in $t$. 

Following similar arguments as in Step 1.4 and Step 2 of the proof of Theorem \ref{thm2}, we can show that $\|\widehat{\Sigma}-\Sigma_0\|_{\infty,\infty}=O_p(n^{-\bar{c}})$ for some constant $\bar{c}>0$. Following similar arguments for \eqref{eqn:thm2step2}, we have that, 
\begin{eqnarray*}
	\begin{split}
		\prob\left( \sqrt{n}\widehat{T}\le t|\mathcal{H}_0 \right) \ge \prob\left( \|N(0,\widehat{\Sigma})\|_{\infty}\le t-2\epsilon_0 \log^{-1/2} n|\widehat{\Sigma} \right)-o(1),\\
		\prob\left( \sqrt{n}\widehat{T}\le t|\mathcal{H}_0 \right) \le \prob\left( \|N(0,\widehat{\Sigma})\|_{\infty}\le t+2\epsilon_0 \log^{-1/2} n|\widehat{\Sigma} \right)+o(1),
	\end{split}	
\end{eqnarray*}
for any sufficiently small $\epsilon_0>0$. Since the little-o terms are uniform in $t\in \mathbb{R}$, we obtain that, 
\begin{eqnarray*}
	&&\sup_t |\prob(\sqrt{n}\widehat{T}\le t|\mathcal{H}_0)-\prob(\|N(0,\widehat{\Sigma})\|_{\infty}\le t|\widehat{\Sigma})|\le o(1)\\
	&+&\sup_t |\prob(\|N(0,\widehat{\Sigma})\|_{\infty}\le t+2\epsilon \log^{-1/2} n|\widehat{\Sigma})-\prob(\|N(0,\widehat{\Sigma})\|_{\infty}\le t-2\epsilon_0 \log^{-1/2} n|\widehat{\Sigma})|.
\end{eqnarray*}
By Theorem 1 of \cite{chernozhukov2017detailed}, the term on the second line can be bounded by $O(1) \epsilon_0 \log^{1/2} B \log^{-1/2} n$, where $O(1)$ denotes some positive constant. Since $B=O(n^c)$, $\log^{1/2} B \log^{-1/2} n=O(1)$. As $\epsilon_0$ grows to zero, this term becomes negligible. Consequently, we obtain that, 
\begin{eqnarray*}
	\sup_t \left| \prob\left( \sqrt{n}\widehat{T}\le t|\mathcal{H}_0 \right) - \prob\left( \|N(0,\widehat{\Sigma})\|_{\infty}\le t|\widehat{\Sigma} \right) \right| \le o(1).
\end{eqnarray*}
As such, the distribution of our test statistic can be well-approximated by that of the bootstrap samples. This completes the proof of Theorem \ref{thm3}. 	
\eop
\bigskip

\subsection{Proof of Theorem \ref{thm4}}
We break the proof into two steps. In Step 1, we show that, under $\mathcal{H}_1^*$, there exist two neural networks functions $f(X)\in \mathbb{H}_1$ and $g(Y)\in \mathbb{H}_2$, such that 
\begin{eqnarray*}
	I(f,g)=\Mean [f(X)-\Mean \{f(X)|Z\}][g(Y)-\Mean \{g(Y)|Z\}] \neq 0,
\end{eqnarray*}
In Step 2, we prove the power of our test approaches one, as the sample size diverges to infinity.

\medskip
\noindent
\textbf{Step 1.} 
We first observe that the measure $I(f,g)=\Mean [f(X)-\Mean \{f(X)|Z\}][g(Y)-\Mean \{g(Y)|Z\}]$ is continuous in $f$ and $g$. That is, for any $f_1,f_2\in L_X^2$ and $g_1,g_2\in L_Y^2$, the difference $I(f_1,g_1)-I(f_2,g_2)$ decays to zero as both $\Mean |f_1(X)-f_2(X)|^2$ and $\Mean |g_1(X)-g_2(X)|^2$ decay to zero. 

Under $\mathcal{H}_1^*$, there exist functions $f^*\in L_X^2$ and $g^*\in L_Y^2$, such that $I(f^*,g^*)\neq 0$. Without loss of generality, assume $f^*$ and $g^*$ are bounded. Otherwise, we can find sequences of bounded functions $\{f_n^*\}_n$ and $\{g_n^*\}_n$ that converge to $f^*$ and $g^*$ under $L_2$-norm, respectively. As a result, we would have $I(f_n^*,g_n^*)\neq 0$ for some $n$.

By Lusin's theorem, we can find a sequence of bounded and continuous functions $\{f_n^{**}\}_n$, such that $\lim_n \prob(f_n^{**}(X)\neq f^*(X))=0$. By dominated convergence theorem, it follows that $f_n^{**}$ converges to $f^*$ under $L_2$-norm. Similarly, we can find a sequence of continuous functions $\{g_n^{**}\}_n$, such that $g_n^{**}$ converges to $g^*$ under $L_2$-norm. This together with the fact that $I(f,g)$ is continuous in $(f,g)$ implies that there exist some continuous functions $f^{**}$ and $g^{**}$, such that $I(f^{**},g^{**}) \neq 0$. 

A key observation here is that, the class of neural networks have universal approximation property. Since the support of $X$ and $Y$ are bounded, it follows from Theorem 1 of \cite{cybenko1989approximation} that the class of single-layered neural networks with sigmoid activation function is dense in the class of bounded, continuous functions with a compact support. As such, we can find some neural network functions $f^{***}$ and $g^{***}$ such that $I(f^{***},g^{***})\neq 0$. We then argue that there must exist $f\in \mathbb{H}_1$ and $g\in \mathbb{H}_2$, such that $I(f,g)=0$. Otherwise, $f^{***}$ and $g^{***}$ can be represented as linear combinations of neural network functions in $\mathbb{H}_1$, $\mathbb{H}_2$ with finitely many number of parameters, and we would have $I(f^{***},g^{***})=0$ as a result. This completes Step 1.

\medskip
\noindent
\textbf{Step 2.} 
We first show that  $I(h_{1,\theta_1},h_{2,\theta_2})$ is a Lipschitz continuous function of $(\theta_1,\theta_2)$. Note that $h_{1,\theta_1}(X)$ and $h_{2,\theta_2}(Y)$ are Lipschitz continuous functions of $\theta_1$ and $\theta_2$, respectively. For any $\theta_{1,1},\theta_{1,2} \in \mathbb{R}^{d_1}$, $\theta_{2,1},\theta_{2,2}\in \mathbb{R}^{d_2}$, we have that, 
\begin{align}
& |I(h_{1,\theta_1},h_{2,\theta_2})-I(h_{1,\theta_1},h_{2,\theta_2})| \nonumber \\ 
\le \; & \left| \Mean [h_{1,1}(X)-\Mean \{h_{1,1}(X)|Z\}-h_{1,2}(X)+\Mean \{h_{2,1}(X)|Z\}][h_{2,1}(Y)-\Mean \{h_{2,1}(Y)|Z\}] \right| \label{secondline} \\ 
& + \left| \Mean [h_{1,2}(X)-\Mean \{h_{1,2}(X)|Z\}][h_{2,1}(Y)-\Mean \{h_{2,1}(Y)|Z\}-h_{2,2}(Y)+\Mean \{h_{2,2}(Y)|Z\}] \right|. \label{thirdline}
\end{align} 
Since the class of functions in $\mathbb{H}_2$ are upper bounded by $O(\sqrt{\log n})$ with probability tending to $1$, the right-hand-side of \eqref{secondline} is bounded from above by
\begin{eqnarray*}
	O(1)\Mean \left| h_{1,1}(X)-\Mean \{h_{1,1}(X)|Z\}-h_{1,2}(X)+\Mean \{h_{2,1}(X)|Z\} \right| \sqrt{\log n},
\end{eqnarray*}
with probability tending to $1$. By Jensen's inequality, the above quantity can be further bounded from above by 
\begin{eqnarray*}
	O(1)\Mean \left| h_{1,1}(X)-h_{1,2}(X) \right| 2 \sqrt{\log n}\le K\|\theta_{1,1}-\theta_{1,2}\|_2\sqrt{\log n},
\end{eqnarray*}
for some constant $K>0$. Following similar arguments, we can show that the right-hand-side of \eqref{thirdline} is bounded from above by $K\|\theta_{2,1}-\theta_{2,2}\|_2\sqrt{\log n}$, for any $\theta_{2,1}$ and $\theta_{2,2}$, with probability tending to $1$. To summarize, conditional on the event that $\mathcal{H}_1$ and $\mathcal{H}_2$ are bounded function classes, we have shown that
\begin{eqnarray*}
	|I(h_{1,\theta_1},h_{2,\theta_2})-I(h_{1,\theta_1},h_{2,\theta_2})| \le K \left( \|\theta_{1,1}-\theta_{1,2}\|_2+\|\theta_{2,1}-\theta_{2,2}\|_2 \right) \sqrt{\log n}.
\end{eqnarray*}
Consequently, for any sufficiently small $\epsilon>0$, there exists a neighborhood $\mathcal{N}=\{(\theta_1,\theta_2):\|\theta_j-\theta_j^*\|_2\le \delta\log^{-1/2} n\}$ for some constant $\delta>0$ around $(\theta_1^*,\theta_2^*)$, such that $I(h_{1,\theta_1},h_{2,\theta_2}) \ge \epsilon$ for any $(\theta_1,\theta_2)$ that belongs to this neighborhood.

Since $(\theta_{1,b},\theta_{2,b})$ are generated from the multivariate normal distribution, and the dimensions $d_1$ and $d_2$ are finite, the probability that $(\theta_{1,b},\theta_{2,b})$ belongs to this neighborhood is strictly greater than $O(\log^{-c_1} n)$ for some constant $c_1>0$. Since $B=c_0 n^c$, the probability that at least one pair of parameters $(\theta_{1,b_1},\theta_{2,b_2})$ belongs to this neighborhood approaches one. Consequently, we have that, 
\begin{eqnarray*}
	\max_{b_1,b_2} \hbox{GCM}^*\left\{ h_{1,b_1}(X),h_{2,b_2}(Y) \right\} \ge \epsilon,
\end{eqnarray*}
with probability tending to $1$. 

Following similar arguments as in the proof of Theorems \ref{thm2} and \ref{thm3}, we can show that $|T-\max_{b_1,b_2} \hbox{GCM}^*\{h_{1,b_1}(X),h_{2,b_2}(Y)\}| = o_p(1)$, and $ \widetilde{T}_j=o_p(1)$. Consequently, both probabilities $\prob(T<\epsilon/2)$ and $\prob(\widetilde{T}_j\ge \epsilon/2)$ converge to zero. Therefore, the probability that the $p$-value is greater than $\alpha$ is bounded by the probability that $\prob(T<\epsilon/2)$, which converges to zero. This completes the proof of Theorem \ref{thm4}.

\bibliography{ref_GANCIT}

\end{document}